\documentclass[Journal,InsideFigs,letterpaper, NoLineNumbers]{ascelike-new}
\usepackage[utf8]{inputenc}
\usepackage[T1]{fontenc}
\usepackage{lmodern}
\usepackage{graphicx}
\usepackage{multirow}
\usepackage[style=base,figurename=Fig.,labelfont=bf,labelsep=period]{caption}
\usepackage{subcaption}
\usepackage{amsmath}
\usepackage{newtxtext,newtxmath}
\usepackage{tabularx}
\newcolumntype{Y}{>{\centering\arraybackslash}X}
\usepackage{booktabs}
\usepackage[colorlinks=true,citecolor=red,linkcolor=black]{hyperref}
\NameTag{Liu, \today}
\begin{document}

\title{Learning Slope-Adaptive Whole-Body Locomotion for Humanoid Robots in Roofing Construction}

\author[1]{Songyang Liu}
\author[2]{Shuai Li}

\affil[1]{Ph.D. Student, Department of Civil and Coastal Engineering, University of Florida, Gainesville, FL 32611. Email: liusongyang@ufl.edu}
\affil[2]{Associate Professor, Department of Civil and Coastal Engineering, University of Florida, Gainesville, FL 32611 (corresponding author). Email: shuai.li@ufl.edu}

\maketitle

\begin{abstract}
Roofing requires workers to coordinate locomotion, balance, and work-related body motions on pitched surfaces, creating a challenging application for humanoid robots. Directly retargeted human demonstrations, however, may preserve motion appearance while placing the robot's feet or hands incorrectly relative to the roof. This study presents a task-semantic scene-grounded framework for learning roofer-style whole-body motions on a Unitree G1. Human demonstrations are captured using a tracking system and retargeted to the robot, while a metric roof model supplies the spatial reference unavailable from the tracking system. A trajectory-level optimization grounds inferred support contacts and annotated work relations to the roof, and execution-aware reinforcement learning encourages the resulting policy to preserve these relations under dynamic tracking errors. The framework is evaluated through a multi-motion tracking study, a roof-pitch coverage matrix, a five-way nailgun ablation, cross-task experiments on hammering and lateral pushing, and comparisons with pure reinforcement learning and zero-shot teleoperation. Our method enables the robot to satisfy support, work-clearance, and nonpenetration criteria across all evaluated seeds. Across nailgun, hammering, and pushing, it achieves work-clearance errors between $0.256$ and $0.531$ cm and $3/3$ successful evaluations per task. Physical experiments reproduce uphill walking, nailgun, hammering, and bending motions with mean base-frame motion errors below $80$ mm. These findings establish scene-grounded human motion learning as a promising basis for construction-oriented humanoid motion primitives.
\end{abstract}

\section{Introduction}
Roofing is one of the most hazardous, labor-intensive, and operationally variable activities in the construction industry. Workers must repeatedly traverse steep or uneven roof surfaces, maintain balance near elevated and often unprotected areas, handle tools and materials, and perform contact-rich work under environmental stressors such as heat, wind, glare, dust, and time pressure. Falls remain the leading cause of death among construction workers~\cite{osha_fall_workbook,cdc_falls_2019}, and roofing is particularly hazardous because work is performed at height and frequently involves sloped surfaces and unstable support conditions~\cite{osha_roofing_workers,bls_fatal_falls_2023,cpwr_falls_2024}. These risks create a strong need for robotic systems that can reduce direct human exposure to dangerous roof work.

Despite this need, roofing remains difficult to automate using current robotic systems. Existing construction robots are often designed for structured, ground-level, or single-function workflows. Roof work, in contrast, requires a robot to access elevated surfaces, stabilize on pitched terrain, transition between locomotion and low working postures, and coordinate body motion with task-oriented actions. Existing roof robotics has made progress in perception, inspection, and task-specific automation, but the broader problem of generating roofer-style whole-body behavior on pitched roofs remains largely unexplored. Roofing is therefore not only an important construction application, but also a challenging frontier problem for embodied robotics.

Humanoid robots are especially relevant to this problem, not because human morphology is universally optimal, but because roofing workflows, access patterns, body postures, tool interfaces, and workspaces are deeply organized around the human body. Roof access routes, kneeling and crouching postures, reach envelopes, hand-tool usage, and material-handling conventions all assume a two-legged, two-armed worker capable of fluidly transitioning between mobility and work. A humanoid robot therefore provides a platform for studying roofing-oriented whole-body autonomy. However, deploying humanoids for roofing is not simply a matter of applying a general locomotion controller to a roof. On a pitched surface, every posture choice affects balance, every local work motion perturbs stance stability, and every step changes the feasible region for subsequent work actions. Specialized wheeled or tracked roofing robots offer lower centers of gravity, higher load capacities, and superior static stability, making them well-suited for repetitive inspection, cleaning, or material handling on roofs with known geometries. While quadrupedal robots provide robust mobility due to their larger support polygons, performing actual tasks usually requires the addition of robotic arms, and they struggle to naturally replicate the bimanual workspace coordination characteristic of human roofing work. Aerial systems excel at inspection and mapping but are limited in terms of payload, endurance, sustained contact, and the ability to exert force. We choose humanoid robots because the tasks discussed here involve not only discrete foothold selection but also human-scale limb extension, postural transitions, and upper-body movements involving tools or materials. A single humanoid platform can switch between states, such as moving, bending, crouching, kneeling, and positioning for work, while adapting to workspaces designed for humans and remaining compatible with existing tools. This versatility, however, does not obscure the inherent drawbacks of bipedal operation: high centers of gravity, limited payload, complex control requirements, and a significant risk of falling when working on inclined surfaces. Therefore, our research focuses on the feasibility of performing various roofing tasks under controlled conditions, rather than asserting that humanoids are superior to specialized roofing platforms in every scenario.

Recent advances in humanoid control have significantly improved the ability of robots to generate agile and human-like whole-body motion. Learning-based locomotion systems have demonstrated robust traversal over challenging natural and man-made terrain~\cite{radosavovic2024challengingterrain,gu2024dwl,sun2025perceptive}. In parallel, large-scale humanoid motion tracking frameworks have shown that diverse human motion priors can be distilled into general controllers capable of tracking broad motion distributions with increasingly human-like behavior~\cite{chen2025gmt,luo2025sonic}. These developments suggest a path toward humanoids that can operate in human-built environments. However, existing capabilities are typically developed and evaluated in settings where walking and working are partially decoupled, the support surface is relatively benign, or task risk is treated as a secondary constraint. Roofing violates these assumptions because locomotion, posture, support stability, and task-oriented motion are tightly coupled on an elevated sloped surface.

\begin{figure}[!htbp]
    \centering
    \includegraphics[width=\linewidth]{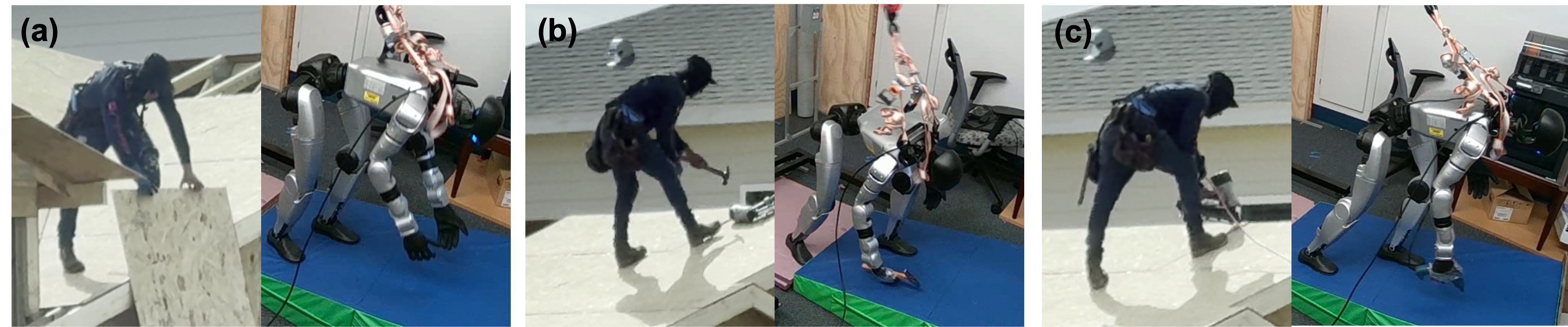}
    \caption{We enable humanoid robots to learn human roofer-style whole-body locomotion skills that transfer from simulation to a physical Unitree G1. Demonstrated behaviors on a slope include (a) bending to reach objects from a low place, (b) hold and brandish a hammer, and (c) hold a nailgun and position it close to the slope surface.}
    \label{fig:1}
    \vspace{-2pt}
\end{figure}

Fig.~\ref{fig:1} illustrates the target capability studied in this work. Roofing operations often require roofers to bend toward a low work area and manipulate a hammer or nailgun near the roof surface. These behaviors require more than stable locomotion on an incline: the robot must preserve the demonstrated posture and coordination while maintaining appropriate support and task-effector placement relative to the roof. The central challenge is therefore to convert human roofing demonstrations into dynamically executable robot motions without losing the spatial relations that make them meaningful for the intended work. 

\begin{figure}
    \centering
\includegraphics[width=\linewidth]{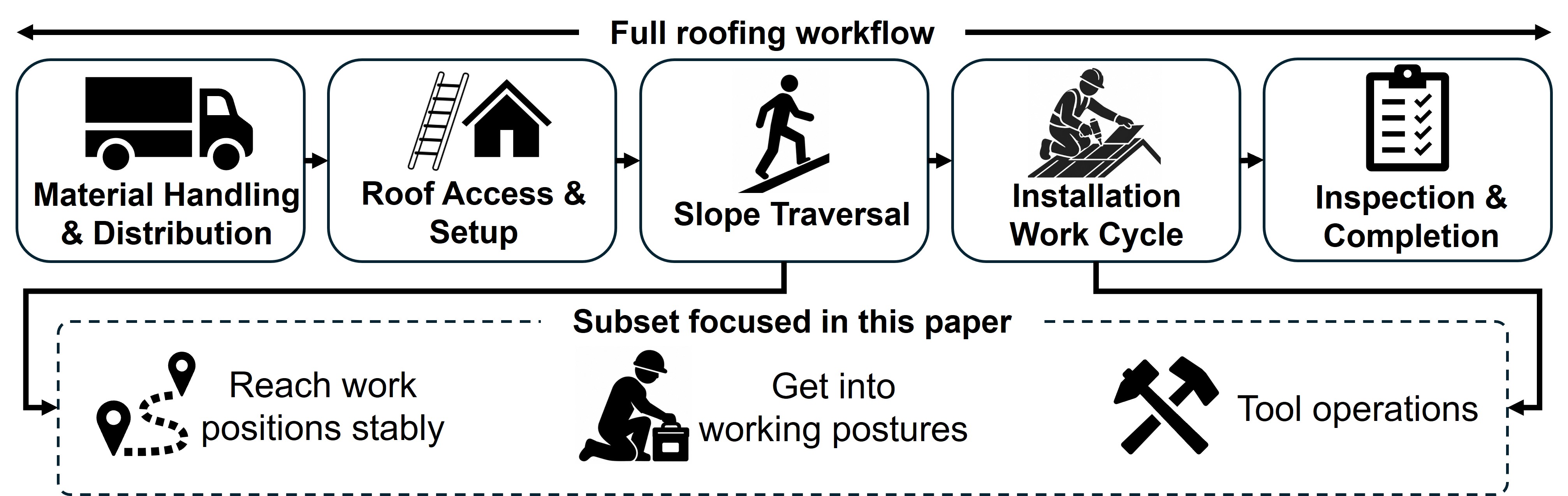}
    \caption{Roofing workflow scope. This paper focuses on slope traversal, working-posture transitions, and representative tool-operation motions.}
    \label{fig:workflow}
\end{figure}

Fig.~\ref{fig:workflow} places these motion capabilities within the broader roofing workflow. Complete roofing automation would require roof access and safety setup, material transportation, slope traversal, installation work cycles, inspection, and quality assessment. This study does not address this full end-to-end process. Instead, it focuses on a motion-centric subset comprising traversal over pitched surfaces, transitions into low working postures, and representative tool- or object-related whole-body motions.

The objective of this study is to learn \emph{slope-adaptive whole-body locomotion} for humanoid robots in roofing construction from human demonstrations. Here, whole-body locomotion encompasses roof traversal, transitions into low working postures, and representative tool- or object-related motions, rather than walking alone. Specifically, we seek to adapt demonstrated roofer-style behaviors to pitched roof geometry while preserving whole-body coordination, maintaining balance and appropriate support, and satisfying task-specific work-clearance and body-surface nonpenetration requirements. This objective couples the construction of geometrically valid reference motions with the learning of feedback policies that preserve the intended motion and task relations during dynamic execution.

Achieving this objective requires addressing a central problem: a human
demonstration may contain appropriate roofer-style motion semantics, while the
retargeted humanoid reference may still be geometrically inconsistent with the
roof. Sparse VR tracking describes how the demonstrator moves and when support
or work occurs, but it does not provide reliable metric
registration between the body and the roof surface. Therefore, a
kinematically plausible reference may place the robot's feet below the roof,
move the hand too far from the work surface, or allow part of the hand mesh to
penetrate the roof. A fixed vertical offset can improve foot placement, but it
moves the complete body and cannot independently preserve a task-specific
hand-surface relation.

To address this problem, we propose a scene-grounded whole-body motion learning framework for the Unitree G1 humanoid. Sparse PICO tracking provides demonstrated motion and task timing, while a metric roof mesh provides the geometry for support and work constraints. We register the retargeted motion to a roof-aligned frame, infer foot-support intervals, and associate each task with a work phase and a desired hand--surface relation. The complete trajectory is then refined using multi-point support, task-effector clearance, body-mesh nonpenetration, motion-preservation, and temporal-smoothness objectives, with planned knee contact included for kneeling motions. Unlike a uniform height offset, this formulation adjusts support and work effectors independently relative to the same roof. Because a feasible reference does not guarantee correct closed-loop execution, we additionally introduce phase-gated clearance and mesh-nonpenetration rewards during reinforcement learning. These rewards preserve the intended hand-surface relation and prevent the hand mesh from penetrating the roof.

The experiments are conducted in two stages. In simulation, we train humanoid policies to track VR-derived roofer motion priors on pitched-roof terrains and evaluate slope coverage across different roof pitches, perform a five-way nailgun ablation covering raw retargeting (A), manual offset (M), support correction (B), reference-level task correction (C), and execution-aware learning (D), and repeat the C-D comparison for hammering and lateral pushing. We also compare our approach with reward-only learning and zero-shot whole-body teleoperation. In the real-world stage, we deploy the trained policies on a physical Unitree G1 and evaluate representative roofing motions using onboard base-frame motion-tracking measurements. These hardware tests provide preliminary deployment evidence rather than demonstrating complete autonomous roofing.

The main contributions of this paper are as follows:

\begin{itemize}
    \item We develop a VR-to-humanoid motion pipeline that converts human roofing demonstrations into retargeted reference motions for humanoid policy learning.

    \item We develop a trajectory-level reference refinement method that
    combines automatically inferred foot-support intervals, task-specific work
    phases, multi-point support anchoring, task-effector clearance,
    body-mesh nonpenetration, and temporal smoothness.

    \item We introduce execution-time task-semantic clearance and mesh-safety
    rewards that encourage the learned policy to preserve the intended
    hand-roof relation despite closed-loop tracking errors.
\end{itemize}

\section{Literature Review}

\subsection{Roof Perception, Robotic Inspection, and Sloped-Surface Mobility}

Roofing has motivated substantial research on automated perception, condition assessment, and robotic inspection because roof inspection is hazardous, time-consuming, and difficult to perform reliably on sloped or elevated surfaces. One major research direction focuses on generating digital representations of roof geometry and roof-surface conditions. \citeN{li2024roofdt} developed an automated pipeline for roof orthophoto generation and semantic segmentation to support digital-twin-based monitoring of slated roofs, while related work further extracted roof sub-components, including slated areas and leadwork, using deep-learning-based semantic segmentation~\cite{li2024roofsubcomponents}. These studies demonstrate the importance of roof geometry, surface layout, and component semantics for automated roof understanding.

Another major direction focuses on detecting and assessing roof defects from aerial or close-range imagery. \citeN{hezaveh2017roofdamage} investigated roof damage assessment using RGB imagery collected by small unmanned aircraft systems and deep learning, while \citeN{alzarrad2022roofaiuav} used unmanned aerial vehicles and deep learning to assess roof conditions and identify missing shingles on sloped roofs. \citeN{xu2022roofdamage} studied roof damage detection and severity classification after Typhoon Faxai using aerial photographs, and \citeN{kucharczyk2025poststormroof} mapped post-storm roof damage with emphasis on roof decking and roof holes. Other studies have addressed more specific maintenance applications. \citeN{zahradnik2023flatroof} proposed UAV-based flat-roof classification and leak detection using RGB and thermal imagery, \citeN{mostafa2023builtuproof} used image analysis to quantify defects and prioritize repairs in built-up roofs, and \citeN{santos2023flatroofequipment} applied deep learning to detect equipment on flat roofs from UAV imagery. More recently, \citeN{zhao2025rrdsegnet} developed a real-time roof-defect segmentation method for robotic inspection, illustrating how perception modules can be incorporated into mobile robotic inspection workflows.

Beyond remote sensing and image-based assessment, roof inspection robots directly address the need to physically access roof surfaces while reducing human exposure to fall hazards. Such systems must operate close to the roof, sense local conditions, and maintain stability under slope-induced mobility constraints, making sloped-surface mobility a central issue in roof robotics. \citeN{zhao2025saha} developed a module-enhanced slope-adaptive and hazard-aware hexapod robotic system for safe roof inspection, explicitly addressing roof-surface hazards and platform adaptation on inclined terrain. This work demonstrates that roof deployment requires more than conventional flat-ground navigation; a robot must regulate its body, maintain support, and avoid hazardous regions while moving on pitched surfaces. Commercial systems are also beginning to integrate multimodal sensing with robotic roof access. For example, autonomous or semi-autonomous platforms such as Roofus combine ground-penetrating radar, thermal cameras, LiDAR, and RGB cameras for roof-condition assessment and inspection planning~\cite{bdr2026roofus}. These developments indicate a progression from image-only documentation toward robotic systems that physically access roof surfaces and collect local multimodal inspection data.

Nevertheless, existing roof-perception and robotic-inspection systems primarily produce roof geometry, semantic labels, damage classifications, or inspection measurements, while the robotic platforms themselves are generally specialized for sensing and mobility. Their objective is to move safely enough to collect inspection data rather than to reproduce the whole-body work patterns of human roofers. Consequently, they do not address humanoid behaviors such as cautious traversal, crouching, kneeling, leaning, reach-oriented positioning, or coordinated locomotion-and-work motions on pitched roofs. Thus, although existing research provides important capabilities in environmental understanding, defect assessment, and sloped-surface robotic access, it does not solve the humanoid whole-body motion-generation problem studied in this paper.

\subsection{Roof Activity Automation}

Beyond inspection and monitoring, a smaller body of work has explored physical automation of roofing activities.~\citeN{romano2021nailedit} demonstrated autonomous roofing with a nailgun-equipped octocopter on an adjustable-slope roof mock-up. Their system used an aerial robot equipped with an off-the-shelf nailgun to fasten shingles, showing that robotic systems can move beyond passive inspection and perform physically interactive roof tasks. This work is important because it demonstrates that roof work can be formulated as a robot execution problem involving roof-frame localization, contact-force application, and task-specific motion planning. Other roof-related automation studies have examined remote or teleoperated robotic systems for hazardous roof operations.~\citeN{ivaldi2023telemovtop} studied teleoperating a robot for removing asbestos tiles on roofs, providing an example of robotic assistance for a roof activity that is dangerous for human workers. This type of work shows that roof automation is not limited to visual inspection; robots may also be used to support hazardous removal, repair, or installation activities. Recent commercial systems further suggest growing interest in robotic shingle installation and roof-work assistance, although these systems are typically designed as specialized tools rather than general humanoid workers~\cite{renovate2026rufus}.

However, existing roof activity automation remains highly task-specific. In the nailgun-equipped octocopter system, the target operation is fastening pre-placed shingles using a specialized aerial platform. In teleoperated roof-tile removal, the target task is remote execution of a hazardous removal operation. These systems do not address how a robot should traverse the roof surface, lower its body into a work posture, maintain balance while leaning, or coordinate locomotion with repeated local roof operations. Thus, although roof activity automation demonstrates the possibility of robotic roof work, it does not solve the broader problem of humanoid roofer locomotion.

Existing roof robotics studies have made progress in roof perception, digital documentation, defect assessment, inspection mobility, and task-specific automation. The remaining gap is not simply detecting roof conditions or executing one specialized roof operation, but learning slope-conditioned, posture-rich, roofer-style whole-body behavior. This motivates our study of humanoid roofer locomotion, where human roofing demonstrations are used as motion priors for learning stable and natural humanoid behaviors on pitched roofs.

\subsection{Humanoid locomotion}

Humanoid locomotion research has developed both model-based and learning-based approaches to maintaining balance and traversing uneven terrain. A representative model-based system is the Atlas framework of \citeN{kuindersma2016atlas}, which integrates optimization-based footstep placement, whole-body planning and control, and state estimation to execute walking plans over non-flat terrain. This work establishes the importance of explicitly accounting for environmental geometry and feasible support regions. In the roofing setting, however, selecting feasible footsteps is only part of the problem: the robot must also maintain support while lowering its body and positioning its hands near an inclined work surface.

Learning-based methods have expanded the range of terrain conditions that humanoids can negotiate without requiring a separately designed controller for each environment. \citeN{radosavovic2024realworld} trained a causal-transformer policy using large-scale reinforcement learning in randomized simulation environments and deployed it on a physical Digit robot without real-world training. By conditioning actions on a history of proprioceptive observations and previous actions, the controller adapts its walking behavior to changing conditions. Similarly, \citeN{gu2024dwl} introduced Denoising World Model Learning, which learns a latent representation from noisy and partially observed sensor histories to reconstruct the robot's state and support robust locomotion. Their experiments demonstrated zero-shot transfer across stairs, inclined ground, snow, and uneven terrain. These results establish a strong basis for terrain-adaptive mobility, but terrain traversal alone does not specify the hand--surface relations required by roofing work.

Perception-driven learning further enables humanoids to anticipate obstacles and regulate foot placement. \citeN{zhuang2025parkour} developed a vision-based whole-body parkour policy that learns multiple obstacle-negotiation behaviors without motion priors, including jumping onto platforms and crossing gaps. Their demonstrations also include overriding arm actions for mobile manipulation, indicating that locomotion policies can support upper-body task execution. Focusing on precise support placement, \citeN{wang2025beamdojo} proposed BeamDojo, which combines a foothold reward accounting for polygonal foot geometry, separate critics for locomotion and foothold rewards, and two-stage reinforcement learning. An onboard LiDAR-based elevation map enables deployment on sparse footholds. This explicit treatment of the foot's spatial extent is particularly relevant to roof support, where a single tracked foot point may not adequately represent the relationship between the sole and the inclined surface.

These studies provide complementary foundations in dynamic balance, adaptation, perception, and support-aware learning. Our focus differs from maximizing traversal capability or obstacle-negotiation agility: we seek to preserve demonstrated roofer-style coordination while simultaneously satisfying foot-support, work-clearance, and body-surface nonpenetration requirements on pitched roofs. Accordingly, roof geometry serves not only as terrain information for locomotion, but also as a metric reference for task-specific whole-body relations. This motivates combining scene-grounded reference refinement with execution-aware motion learning, rather than relying on terrain-adaptive walking alone.

\subsection{Human-to-Humanoid Retargeting and
Reference-Conditioned Control}

Reference-conditioned reinforcement learning enables a simulated or
physical humanoid to reproduce motion clips while retaining robustness to
disturbances. DeepMimic established the combination of motion-imitation and
task objectives for learning physics-based character skills
\cite{peng2018deepmimic}, while PHC extended motion imitation to large motion
collections and fail-state recovery \cite{luo2023perpetual}. Recent systems have
transferred reference-conditioned whole-body controllers to physical
humanoids. OmniH2O uses kinematic human pose as a control interface and
learns deployable policies through privileged-to-sparse observation
distillation \cite{he2024omnih2o}, whereas ASAP explicitly learns a residual
model to reduce the simulation-to-real dynamics mismatch
\cite{he2025asap}. BeyondMimic combines robust motion tracking with a unified diffusion policy for test-time control and composition of learned motion skills~\cite{liao2026beyondmimic}. These approaches improve motion acquisition, tracking, or dynamics transfer, whereas our study focuses on phase-dependent support and work-clearance relations with a measured roof surface.

A separate line of research studies the quality of human-to-humanoid motion
retargeting. GMR shows that foot sliding, self-penetration, and infeasible
kinematics in retargeted references can substantially affect downstream
policy robustness \cite{araujo2025retargeting}. OmniRetarget further preserves
agent--terrain and agent--object interactions through an interaction-mesh
representation~\cite{yang2025omniretarget}. These studies provide the
retargeting and tracking foundations used by our pipeline. Our contribution
is not a new general-purpose retargeter or tracking architecture. Instead,
we refine an existing robot reference relative to a metric roof model and
explicitly represent phase-dependent support and noncontact work relations
that must also be preserved during policy execution.

\section{Methodology}
\begin{figure}
    \centering
    \includegraphics[width=1\linewidth]{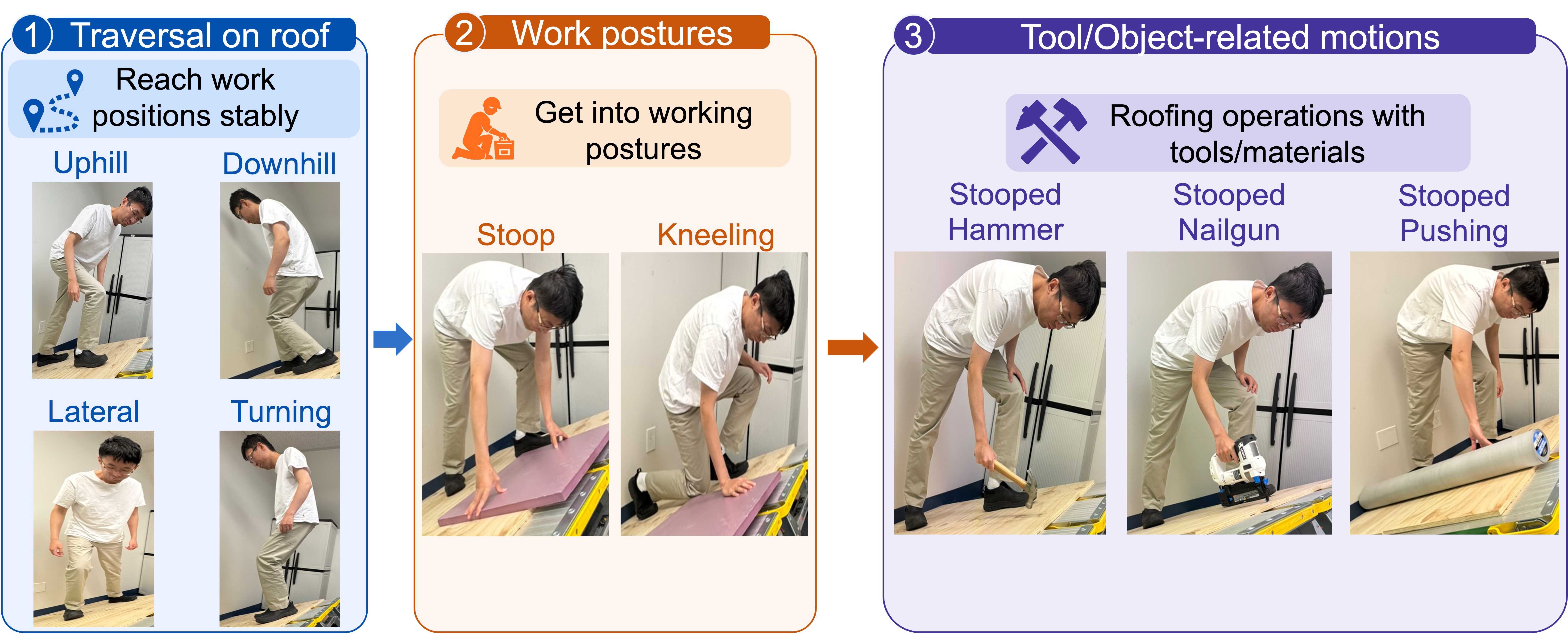}
    \caption{Roofing-oriented motion taxonomy. The studied skills include roof traversal, work postures, and representative tool/object-related motions.}
    \label{fig:motion_taxonomy}
\end{figure}

Our goal is to enable a humanoid robot to acquire \emph{roofer-style} whole-body behaviors on pitched roofs. Unlike generic slope locomotion, the target behaviors in this work include both movement across the roof surface and posture-rich work motions near the roof surface. Fig.~\ref{fig:motion_taxonomy} summarizes the roofing-oriented motion families studied in this paper. We organize the target behaviors into three levels. The first level focuses on roof traversal, including uphill, downhill, lateral, and pivoting motions for reaching nearby work locations. The second level focuses on work postures, such as stooping and kneeling, which allow the body to approach the roof surface. The third level includes tool- or object-related motions, such as stooped hammering, nailgun use, and material pushing. This taxonomy defines the scope of the human demonstrations collected in this work and provides the motion priors used for humanoid policy learning.

\begin{figure}
    \centering
    \includegraphics[width=1\linewidth]{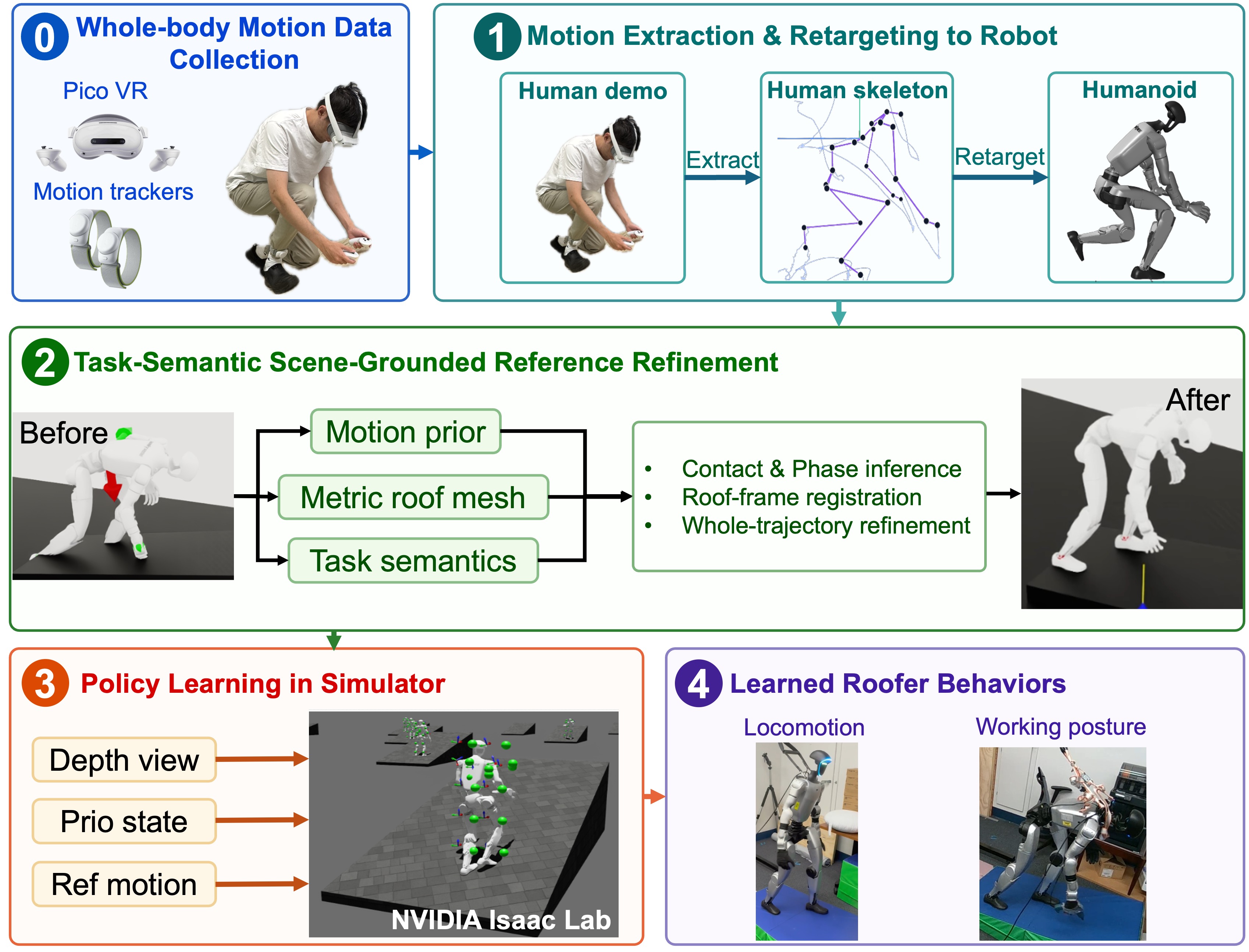}
    \caption{Overview of the proposed humanoid roofer skill learning framework, including VR-based motion collection, humanoid retargeting, task-semantic scene-grounded reference refinement, and policy learning in simulation.}
    \label{fig:pipeline}
\end{figure}

Fig.~\ref{fig:pipeline} illustrates the proposed scene-grounded whole-body motion learning framework. First, sparse human roofing demonstrations are collected using a PICO headset, hand controllers, and ankle trackers to provide motion trajectories and temporal task semantics. The captured motion is then reconstructed and retargeted to the Unitree G1 embodiment to obtain an initial robot motion prior. Next, the motion prior, a metric roof mesh, and task-specific information, including work phases, desired hand-surface distances, and planned support contacts are jointly used for scene-grounded reference refinement. This stage registers the motion in a roof-aligned frame, infers support and contact phases, and performs whole-trajectory optimization using multi-point support anchoring, task-effector clearance, body-mesh nonpenetration, motion-preservation, and temporal-smoothness objectives. The refined reference and semantic phase masks are subsequently used to train a whole-body tracking policy in simulation. In addition to standard motion-tracking objectives, phase-gated task-clearance and mesh-nonpenetration rewards encourage the executed motion to preserve the intended roofing interaction. Finally, the learned policies are evaluated through simulation ablations and physical-G1 experiments using support, work-clearance, collision, task-success, and motion-tracking metrics.

\subsection{Problem Formulation}

We study the problem of learning scene-grounded whole-body behaviors for a
humanoid robot operating on pitched roof surfaces. In addition to maintaining
balance and tracking human-like motion, the robot must satisfy task-dependent
spatial relations with the roof, such as stable foot support, planned knee
contact, a prescribed hand-surface distance, and body-mesh nonpenetration.
Accordingly, the problem consists of two coupled stages: constructing a
geometrically valid reference motion and learning a feedback policy that
preserves the intended motion and task relations during dynamic execution.

Let
\[
\bar{\mathcal{M}}
=
\left\{
\bar{\mathbf q}_t,
\dot{\bar{\mathbf q}}_t,
\bar{\mathbf p}^{\,b}_t,
\bar{\mathbf R}^{\,b}_t
\right\}_{t=1}^{T}
\]
denote the initial G1 motion prior obtained by retargeting a sparse VR
demonstration, and let \(\mathcal{G}\) denote the metric roof mesh registered
to the motion coordinate frame. The task-semantic specification is written as
\[
\Psi =
\left\{
c^{\mathrm{sup}}_{t,k},
c^{\mathrm{work}}_t,
c^{\mathrm{knee}}_{t,k},
d_w
\right\}_{t=1}^{T},
\]
where \(c^{\mathrm{sup}}_{t,k}\) identifies the support state of contact point
\(k\), \(c^{\mathrm{work}}_t\) identifies the task-relevant work phase,
\(c^{\mathrm{knee}}_{t,k}\) represents planned knee contact when applicable,
and \(d_w\) is the desired task-effector clearance from the roof. The
scene-grounded reference is obtained through whole-trajectory refinement:
\begin{equation}
\label{eq:reference_refinement}
\mathcal{M}^{*}
=
\arg\min_{\mathcal{M}}
\lambda_{\mathrm{sup}}\mathcal{L}_{\mathrm{sup}}
+
\lambda_{\mathrm{work}}\mathcal{L}_{\mathrm{work}}
+
\lambda_{\mathrm{mesh}}\mathcal{L}_{\mathrm{mesh}}
+
\lambda_{\mathrm{prior}}\mathcal{L}_{\mathrm{prior}}
+
\lambda_{\mathrm{smooth}}\mathcal{L}_{\mathrm{smooth}} .
\end{equation}

These terms respectively enforce planned support on the roof, task-effector
clearance, body-mesh nonpenetration, preservation of the retargeted motion
semantics, and temporal smoothness. This formulation permits the lower-body
configuration to adapt to the roof geometry while retaining the characteristic
upper-body structure of the demonstrated roofing behavior.

Given the refined reference \(\mathcal{M}^{*}\), policy learning is formulated
as a reference-conditioned partially observable Markov decision process
\[
\mathcal{P}
=
(\mathcal{S},\mathcal{A},\mathcal{O},P,r,\gamma).
\]
The simulator state \(s_t\in\mathcal{S}\) contains the full robot state,
root state, contact state, reference phase, and terrain configuration. The
29-dimensional action \(\mathbf a_t\in\mathcal{A}\) specifies joint-position
offsets that are converted into desired joint positions and tracked by joint
PD controllers. The actor receives
\[
\mathbf o_t =
\left[
\mathbf c^{\mathrm{ref}}_t,\,
\mathbf R^{\mathrm{ref},b}_t,\,
\boldsymbol{\omega}^{b}_t,\,
\mathbf q_t-\mathbf q_0,\,
\dot{\mathbf q}_t,\,
\mathbf a_{t-1}
\right],
\]
where \(\mathbf c^{\mathrm{ref}}_t\) contains the reference joint positions
and velocities, \(\mathbf R^{\mathrm{ref},b}_t\) contains reference body
orientations expressed in the robot base frame,
\(\boldsymbol{\omega}^{b}_t\) is the base angular velocity, and
\(\mathbf q_t-\mathbf q_0\), \(\dot{\mathbf q}_t\), and
\(\mathbf a_{t-1}\) are the current joint-position offsets, joint velocities,
and previous action, respectively. This vector summarizes the reference and current proprioceptive components. For the perceptive configuration illustrated in Fig.~\ref{fig:flow}, the actor additionally receives projected gravity, proprioceptive history, and a depth observation. The depth observation is rendered with noise in simulation and supplied by the onboard camera during deployment. The registered roof mesh and semantic phase masks are used for reference refinement and training rewards, not supplied directly as actor observations.

The policy \(\pi_{\theta}(\mathbf a_t\mid\mathbf o_t)\) is trained to maximize
\[
J(\theta)
=
\mathbb{E}_{\pi_{\theta}}
\left[
\sum_{t=0}^{T-1}\gamma^t r_t
\right],
\]
with
\[
r_t =
r^{\mathrm{WBT}}_t
+
\lambda_{\mathrm{clear}}r^{\mathrm{clear}}_t
-
\lambda_{\mathrm{pen}}c^{\mathrm{pen}}_t
+
\lambda_{\mathrm{contact}}r^{\mathrm{contact}}_t
-
\lambda_{\mathrm{reg}}c^{\mathrm{reg}}_t .
\]
Here, \(r^{\mathrm{WBT}}_t\) contains the standard whole-body pose,
orientation, and velocity tracking objectives;
\(r^{\mathrm{clear}}_t\) preserves the desired task-effector clearance during
the work phase; \(c^{\mathrm{pen}}_t\) penalizes body-roof penetration;
\(r^{\mathrm{contact}}_t\) handles planned contacts such as kneeling; and
\(c^{\mathrm{reg}}_t\) contains action-rate, joint-limit, and undesired-contact
regularization. The roof mesh and semantic phase masks are therefore used for
offline reference construction and training-time reward computation, but are
not required as online actor observations.
\subsection{Whole-body Human Motion Data Collection}

\begin{figure}
    \centering
    \includegraphics[width=1\linewidth]{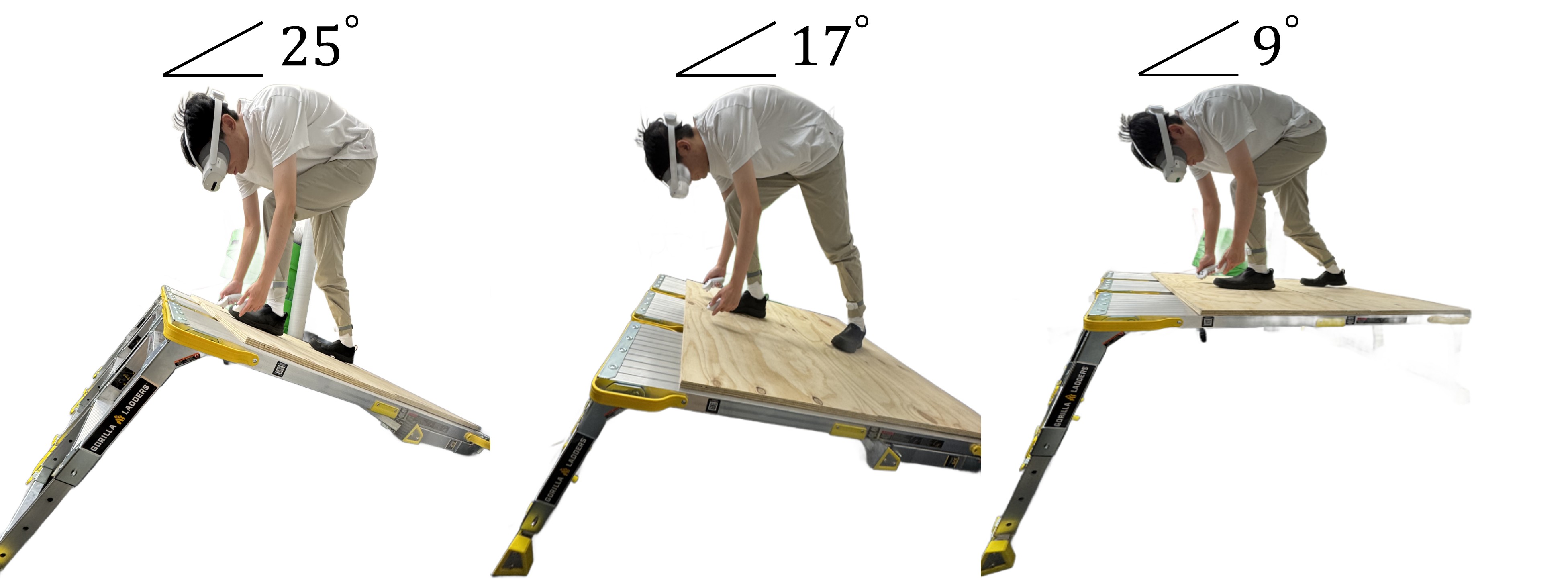}
    \caption{Human demonstration collection on an angle-adjustable sloped-roof platform.}
    \label{fig:platform}
\end{figure}

To provide task-specific motion priors for roofer-style humanoid learning, we collect human demonstrations on an angle-adjustable sloped-roof platform, as shown in Fig.~\ref{fig:platform}. The platform is constructed using wooden plates mounted on flexible ladders, allowing us to emulate pitched roof surfaces under controlled laboratory conditions. In this study, we collect demonstrations on three roof inclinations, \(9^\circ\), \(17^\circ\), and \(25^\circ\), which approximately correspond to commonly referenced residential roof pitch ranges from 2:12 to 6:12~\cite{iko2023minimumslope}. These angles allow us to record roofing-relevant whole-body behaviors under mild-to-steeper sloped conditions. 

The demonstrations are recorded using a VR-based motion-capture setup. The setup consists of a Pico headset, two hand controllers, and two motion trackers attached near the ankles. This configuration is chosen to capture the key motion components required for roofing-relevant behaviors, including head and torso motion, hand trajectories, and lower-limb movement during slope traversal and low-posture transitions. Each demonstration is recorded as a time-indexed text log. Each line in the log corresponds to one tracking frame and contains a nanosecond timestamp, the headset pose, the left and right controller poses, and the ankle tracker poses. Each pose is represented by a 3D position and a quaternion orientation. In addition to these raw device poses, the Pico tracking stream provides a reconstructed full-body motion sequence containing 24 human body joints at the native recording rate.

This study uses human demonstrations as motion priors for engineering validation of the robot-learning framework rather than as a population-level human-behavior dataset. All demonstrations were performed by one adult laboratory researcher with a body height of $1.84$ m and without professional roofing experience. The demonstrator rehearsed each motion on the laboratory platform before recording. Age and other demographic attributes were not collected because the study did not investigate inter-participant or human-behavior effects.

The motion data comprise two separately recorded collections. The multi-slope motion library contains nine motion types: uphill walking, downhill walking, lateral walking, pivoting, stooping, kneeling, hammering, nailgun positioning, and pushing. Each motion was demonstrated three times at each of the $9^\circ$, $17^\circ$, and $25^\circ$ inclinations, yielding 81 sequences. After retargeting and resampling to $30$ Hz, these sequences contain 356--809 frames and last $11.87$--$26.97$ s, with a total of $42{,}349$ robot-reference frames and approximately $1{,}411.63$ s ($23.53$ min). These frame counts describe the $30$-Hz retargeted library, not the native VR logs or the subsequent $50$-Hz policy references.

The separate task-semantic collection contains ten retained raw recordings, with individual durations of $4.26$--$14.96$ s and a total duration of approximately $110.96$ s. One retained take was used for each recorded task instance in this collection; repeated demonstrations were not pooled. Five of these raw recordings are individually documented here: nailgun positioning contains 822 frames over $11.41$ s; hammering, 971 frames over $13.48$ s; pushing, 1,184 frames over $13.17$ s; bending, 842 frames over $9.35$ s; and kneeling, 944 frames over $10.48$ s. These five recordings are a subset of the ten-recording collection, not an additional dataset. Native PICO sampling rates range from approximately $72$ to $90$ Hz.

The PICO system records timestamped head, controller, and body-tracker poses. The signals are resampled to $30$ Hz for humanoid retargeting, and processed robot references are subsequently represented at $50$ Hz for policy training. These rates distinguish native VR acquisition, retargeting, and policy-reference playback. No subject-level training/test split is used because there is only one demonstrator.

Statistical evaluation is performed over robot training seeds and rollout episodes rather than over human demonstrators. The results therefore evaluate whether motion supplied by one demonstrator can serve as an effective robot motion prior; they should not be interpreted as conclusions about general human roofing behavior. The present study does not evaluate inter-person motion variability or generalization of preprocessing across body sizes and individual work styles.

These dataset statistics describe retained recordings. Counts of unsuccessful recording attempts and excluded takes are not reported; the retained sequence counts should not be interpreted as evidence that no trials were excluded.

\subsection{Motion Retargeting to the Humanoid Embodiment}

The captured human motion cannot be directly used as a humanoid reference because the human body and the Unitree G1 differ in kinematic topology, limb proportions, joint limits, and available degrees of freedom. We therefore retarget each human demonstration to the G1 embodiment before using it for policy training.

Let
\[
\mathcal{M}^{h}=\{(\mathbf{x}^{h}_{t,i}, \mathbf{R}^{h}_{t,i})\}_{t=1,i=1}^{T,N_h}
\]
denote the captured human body sequence, where \(\mathbf{x}^{h}_{t,i}\) and \(\mathbf{R}^{h}_{t,i}\) are the position and orientation of the \(i\)-th tracked human body joint at time \(t\). In our implementation, the Pico body stream provides a 24-joint full-body representation. We select the joints required by the retargeting model, including pelvis, torso, hips, knees, feet, shoulders, elbows, and wrists, and map them to the XRobot human keypoint convention used by General Motion Retargeting (GMR)~\cite{araujo2025retargeting}.

The retargeting process solves an inverse-kinematics matching problem from the selected human keypoints to the Unitree G1 model. The target robot motion is represented as
\[
\mathcal{M}^{r}=\{(\mathbf{q}^{r}_t,\mathbf{p}^{r}_t,\mathbf{R}^{r}_t)\}_{t=1}^{T},
\]
where \(\mathbf{q}^{r}_t\) is the 29-DoF G1 joint configuration, and \(\mathbf{p}^{r}_t,\mathbf{R}^{r}_t\) are the robot root position and orientation. The IK objective matches semantically corresponding body parts between the human and robot, such as pelvis-to-base, torso-to-torso, feet-to-feet, and wrists-to-wrists, while respecting the robot kinematic chain.

After the retargeting, the resulting humanoid motion is converted into the Isaac Lab motion format using the G1 URDF model. We use the pelvis frame as the source reference frame and the G1 \texttt{torso\_link} as the target root frame. The final motion file stores the retargeted root trajectory, root quaternion, joint positions, joint names, and framerate, making it directly replayable as a reference motion in Isaac Lab.

\subsection{Task-Semantic Scene-Grounded Reference Refinement}
\label{sec:scene_grounded_refinement}

This subsection specifies how the terms in
Eq.~\eqref{eq:reference_refinement} are constructed. The symbols
\(\bar{\mathcal M}\), \(\mathcal G\), and \(\Psi\) respectively denote the
initially retargeted G1 motion, metric roof mesh, and task-semantic
specification defined in the preceding subsection. The refinement modifies
the robot reference while keeping the roof geometry fixed.

The manual inputs are the measured initial placement relative to the roof, the annotated work phase, and the task-specific target clearance. The target hand-center clearances are $25$ cm for nailgun positioning, $12$ cm for hammering, and $18.5$ cm for lateral pushing. After roof-frame registration, support candidates are inferred automatically using sole-center speed and acceleration, followed by gap filling and minimum-duration filtering. Support segments are assigned to the floor or roof from the sole-center location within the roof footprint, and five sole points per foot are anchored to the assigned surface. Whole-trajectory optimization then adjusts support and work-effectors jointly with mesh-nonpenetration, motion-preservation, and temporal-smoothness terms. The proposed refinement does not use frame-by-frame manual height editing; a manually selected uniform vertical offset is retained only as baseline M.

\subsubsection{Roof-frame registration}
We first register the retargeted motion and the roof mesh in a common metric
frame. For the selected roof patch, we define the roof-aligned basis
\[
\mathbf R_{\mathcal G}
=
\begin{bmatrix}
\mathbf t_u & \mathbf t_v & \mathbf n
\end{bmatrix},
\]
where \(\mathbf t_u\) is the uphill tangent, \(\mathbf t_v\) is the
cross-slope tangent, and \(\mathbf n\) is the outward roof normal. The rigid
transformation \(\mathbf T_{\mathcal G\leftarrow\mathcal M}\) is determined
using the measured initial toe-to-edge distance and the lateral relation
between the foot midpoint and the roof centerline. Thus, the motion is placed
relative to the roof using physically measured quantities rather than a
manually selected world-\(z\) translation.

For a point \(\mathbf p\), let
\(\Pi_{\mathcal G}(\mathbf p)\) denote its closest point on the selected roof
patch. Its signed roof-normal gap is
\[
d_{\mathcal G}(\mathbf p)
=
\mathbf n^{\top}
\left(
\mathbf p-\Pi_{\mathcal G}(\mathbf p)
\right),
\]
where \(d_{\mathcal G}(\mathbf p)<0\) indicates penetration into the roof.

\subsubsection{Foot-support inference}
The foot-support masks in \(\Psi\) are inferred from the retargeted robot
motion. For each foot \(f\), we smooth its sole-center trajectory and compute
the translational speed and acceleration:
\[
v_{t,f}
=
\left\|
\frac{\partial \mathbf p_{t,f}}{\partial t}
\right\|_2,
\qquad
a_{t,f}
=
\left\|
\frac{\partial^2 \mathbf p_{t,f}}{\partial t^2}
\right\|_2 .
\]
A frame is marked as a support candidate when
\[
c^{\mathrm{sup}}_{t,f}
=
\mathbb I
\left[
v_{t,f}<\tau_v
\land
a_{t,f}<\tau_a
\right].
\]
Short interruptions are filled, and intervals shorter than a prescribed
minimum duration are removed. Each support segment is assigned to either the
flat floor or the roof according to whether its smoothed sole center lies
inside the roof footprint.

Each G1 sole is represented by five contact points. For a support segment
\(\mathcal I_{f,s}\), we calculate the median position of each point over the
segment and project it onto the assigned surface. This produces a fixed
support anchor \(\mathbf p^{*}_{f,j,s}\), where \(j\) indexes the sole point.
The foot component of the support objective is
\[
\mathcal L_{\mathrm{foot}}
=
\sum_f
\sum_s
\sum_{t\in\mathcal I_{f,s}}
\sum_{j=1}^{5}
\left\|
\mathbf p_{f,j}
\left(
\mathbf q_t,\mathbf p^{r}_t
\right)
-
\mathbf p^{*}_{f,j,s}
\right\|_2^2 .
\]
Using multiple sole points prevents a solution in which the ankle origin is
above the roof while another part of the foot penetrates the surface.

\subsubsection{Task-semantic work relation}
The work-phase mask \(c^{\mathrm{work}}_t\) identifies the interval in which
the task effector must maintain a prescribed relation with the roof. It is
obtained from the synchronized VR timeline and task annotation rather than
from foot-support detection. The desired clearance \(d_w\) is measured for
the corresponding human operation and represents the normal distance from the
hand center to the work surface.

For each work-phase frame, the target hand position is
\[
\mathbf p^{*}_{h,t}
=
\Pi_{\mathcal G}
\left(
\bar{\mathbf p}_{h,t}
\right)
+
d_w\mathbf n .
\]
This construction retains the tangential hand location from the demonstrated
operation while replacing its potentially inaccurate absolute height with the
metric roof geometry. The corresponding objective is
\[
\mathcal L_{\mathrm{work}}
=
\sum_{t=1}^{T}
c^{\mathrm{work}}_t
\left\|
\mathbf p_h
\left(
\mathbf q_t,\mathbf p^{r}_t
\right)
-
\mathbf p^{*}_{h,t}
\right\|_2^2 .
\]
For behaviors without an intended hand--surface relation, such as the
bending-only and kneeling motions considered in this study,
\(c^{\mathrm{work}}_t=0\); their hands are governed only by the safety
constraint described below.

\subsubsection{Body-mesh nonpenetration}
A correct hand-center distance does not ensure that the complete hand geometry
remains outside the roof. Let \(\mathcal V_b\) contain the vertices of the
collision hull associated with a safety-critical body \(b\), and let
\(\mathbf p_{b,v,t}\) be the world position of vertex \(v\). We define
\[
\mathcal L_{\mathrm{mesh}}
=
\sum_t
c^{\mathrm{safe}}_t
\sum_{b\in\mathcal B_{\mathrm{safe}}}
\left[
m_b-
\min_{v\in\mathcal V_b}
d_{\mathcal G}
\left(
\mathbf p_{b,v,t}
\right)
\right]_+^2,
\]
where \([x]_+=\max(0,x)\), \(m_b\) is a small clearance margin, and
\(c^{\mathrm{safe}}_t\) activates the constraint over the relevant approach,
work, and retraction interval. The lowest collision-hull vertex is
re-evaluated during optimization so that the constraint remains valid as the
hand or knee rotates. Tool-use and pushing motions apply this term to the task
hand, whereas the bending-only motion applies it to both hands.

\subsubsection{Planned knee contact}
Kneeling requires intentional contact that should be distinguished from
accidental penetration. We represent each knee by a proximal collision patch
and examine its roof-normal gap and velocity. A planned knee-contact interval
is inferred when the patch remains close to the roof, moves below a velocity
threshold, and satisfies a minimum-duration requirement. The corresponding
target is obtained by projecting the active knee patch onto the roof with a
small contact margin.

Let \(\mathcal I^{\mathrm{knee}}_{f}\) denote the inferred contact interval for
knee \(f\). Its contact objective is
\[
\mathcal L_{\mathrm{knee}}
=
\sum_f
\sum_{t\in\mathcal I^{\mathrm{knee}}_{f}}
\left\|
\mathbf p^{\mathrm{knee}}_{t,f}
-
\mathbf p^{*,\mathrm{knee}}_{t,f}
\right\|_2^2 .
\]
The support term in Eq.~\eqref{eq:reference_refinement} is consequently
implemented as
\[
\mathcal L_{\mathrm{sup}}
=
\mathcal L_{\mathrm{foot}}
+
\lambda_{\mathrm{knee}}\mathcal L_{\mathrm{knee}},
\]
with \(\mathcal L_{\mathrm{knee}}=0\) for motions without planned knee
contact. Outside the inferred contact interval, the knee remains subject to
the mesh-nonpenetration constraint.

\subsubsection{Trajectory-level parameterization and regularization}
The optimization variables at frame \(t\) are
\[
\boldsymbol{\delta}_t
=
\begin{bmatrix}
\delta r_{n,t} &
\delta\mathbf q_t^{\top}
\end{bmatrix}^{\top},
\]
where \(\delta r_{n,t}\) changes the root position only along the roof normal
and \(\delta\mathbf q_t\) modifies a selected subset of leg, waist, and
task-arm joints. The refined configuration is
\[
\mathbf p^{r,*}_t
=
\bar{\mathbf p}^{r}_t
+
\delta r_{n,t}\mathbf n,
\qquad
\mathbf q^{*}_t
=
\bar{\mathbf q}_t
+
\mathbf E\delta\mathbf q_t,
\]
where \(\mathbf E\) maps the optimized joint subset into the complete robot
configuration.

Deviation from the retargeted motion is regularized by
\[
\mathcal L_{\mathrm{prior}}
=
\sum_{t=1}^{T}
\left\|
\boldsymbol{\delta}_t
\right\|_{\mathbf W}^{2}.
\]
Temporal consistency is enforced using
\[
\mathcal L_{\mathrm{smooth}}
=
\lambda_{\mathrm{vel}}
\sum_{t=2}^{T}
\left\|
\boldsymbol{\delta}_t-\boldsymbol{\delta}_{t-1}
\right\|_2^2
+
\lambda_{\mathrm{acc}}
\sum_{t=2}^{T-1}
\left\|
\boldsymbol{\delta}_{t+1}
-2\boldsymbol{\delta}_t
+\boldsymbol{\delta}_{t-1}
\right\|_2^2 .
\]
The first-frame correction is additionally regularized to avoid an abrupt
transition from the deployment initialization pose, and all optimized joints
are bounded by the G1 joint limits. The resulting reference
\(\mathcal M^{*}\) and its support, work, safety, and planned-contact masks are
then passed to the policy-learning stage described in the following
subsection.
\subsection{Humanoid Learning in Simulation}

We train the humanoid policy in Isaac Lab using robot motion tracking formulation~\cite{zhuang2026deep,yang2025omniretarget}. Each training episode samples a retargeted reference motion together with its paired terrain from the motion metadata file. The terrain mesh is kept fixed in the scene, and the robot is initialized near the corresponding reference state with small reset perturbations. The policy outputs joint-position targets for the 29-DoF Unitree G1 model, which are then tracked by the low-level actuator model.

\begin{figure}
    \centering
    \includegraphics[width=\linewidth]{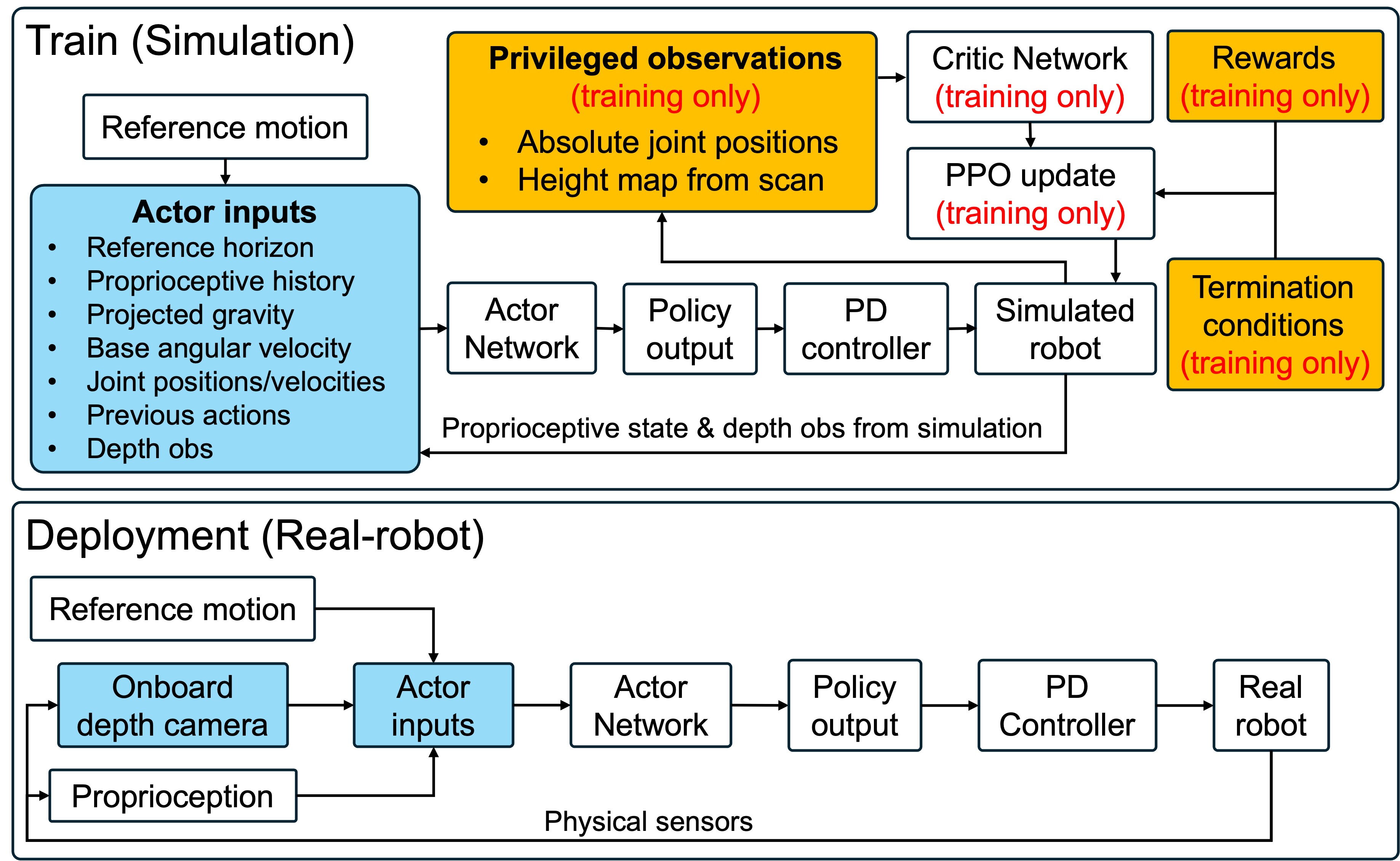}
    \caption{Training and deployment information flow of the asymmetric actor-critic policy. Privileged observations are available only to the critic during training, whereas the deployed actor uses reference motion, onboard depth, and proprioception to generate PD-controlled whole-body motion.}
    \label{fig:flow}
\end{figure}

At each control step, the policy receives both reference-motion information and robot sensory feedback. The reference input includes future joint-position and joint-velocity commands, as well as relative root-position and root-orientation commands derived from the retargeted demonstration. The motion reference is updated at 50 Hz and provides a short reference horizon of 10 frames with 0.1 s spacing. The policy observation also includes proprioceptive history, including projected gravity, base angular velocity, relative joint positions, relative joint velocities, and the previous action. For perceptive training, the policy additionally receives a noisy depth image rendered from a torso-mounted camera. The critic receives privileged observations, including link positions, link orientations, height scan, and proprioceptive state. Fig.~\ref{fig:flow} shows the information flow of the asymmetric actor-critic policy.

The actor output is converted into joint-position targets tracked by PD controllers. In simulation, the resulting robot state supplies feedback to the actor, privileged state to the critic, and the executed configuration used to calculate motion-tracking, task-clearance, mesh-safety, and regularization rewards. Reference exhaustion, terrain-boundary violations, and excessive deviations in root height, projected gravity, or key-link height determine episode termination. The critic, reward calculation, and PPO updates are training-only components. During deployment, the actor and PD-control path remain active with the reference sequence and onboard observations; the safety hoist and operator-controlled emergency stop are separate physical safeguards, not learned termination guarantees.

We do not rely on a separate hand-designed forward-progress reward. Instead, progression along the roof is induced by tracking the retargeted reference motion itself. Training episodes terminate when the reference motion is exhausted, when the robot moves outside the terrain boundary, or when the robot deviates too far from the reference in root height, projected gravity, or key robot link height. These termination conditions prevent failed rollouts from dominating training and encourage the policy to remain close to the reference motion. To improve robustness, we also use domain randomization, including humanoid contact-material properties, actuator gains, selected rigid-body masses, torso center of mass, camera pose offsets, and randomized reset states.

\section{Experiments and Results}

\subsection{Experimental Setup}

The evaluation is organized into simulation and physical-robot stages. In simulation, we first assess tracking performance across the complete roofer-motion library and examine how training-slope coverage affects execution at different roof pitches. We then use a controlled nailgun ablation to isolate the contributions of support grounding, task-semantic reference refinement, and execution-time enforcement, before validating the method on hammering and lateral pushing. Reward-only RL and zero-shot SONIC teleoperation are additionally evaluated as diagnostic baselines. In the physical stage, selected simulation-trained policies are deployed on the Unitree G1 without real-world policy optimization, and onboard robot states are recorded to quantify base-frame motion fidelity.

Two policy-training protocols are used. In the motion-coverage and slope-coverage experiments, one multi-motion policy is trained for each specified slope set: $17^\circ$, $9^\circ+17^\circ$, or $9^\circ+17^\circ+25^\circ$. Each policy learns the motion types represented in its training library; it is not a separate policy for every motion type. The slope matrix evaluates uphill walking at each target pitch. At a training pitch, the evaluation reference belongs to that policy's training library. At an unseen pitch, the corresponding slope-specific reference and terrain were excluded from training. This protocol tests coverage of discrete roof pitches, not a held-out-demonstrator or held-out-motion-type split.

For the scene-grounding experiments, policies are trained separately for each task, ablation condition, and seed. The nailgun A/M/B/C/D conditions therefore use independently trained policies and their respective reference variants, while hammering and pushing each use separate C and D policies. Within a condition, training and evaluation use the same reference clip. These experiments measure tracking and preservation of scene-relative constraints across training seeds, not generalization to unseen human motions. The principal ablation and cross-task comparisons use seeds 42, 123, and 456. Training includes robot-state, contact, friction, action-delay, and external-disturbance randomization; deterministic evaluation disables pose noise, external pushes, action delay, center-of-mass perturbations, and friction variation.

\begin{figure}
    \centering
    \includegraphics[width=\linewidth]{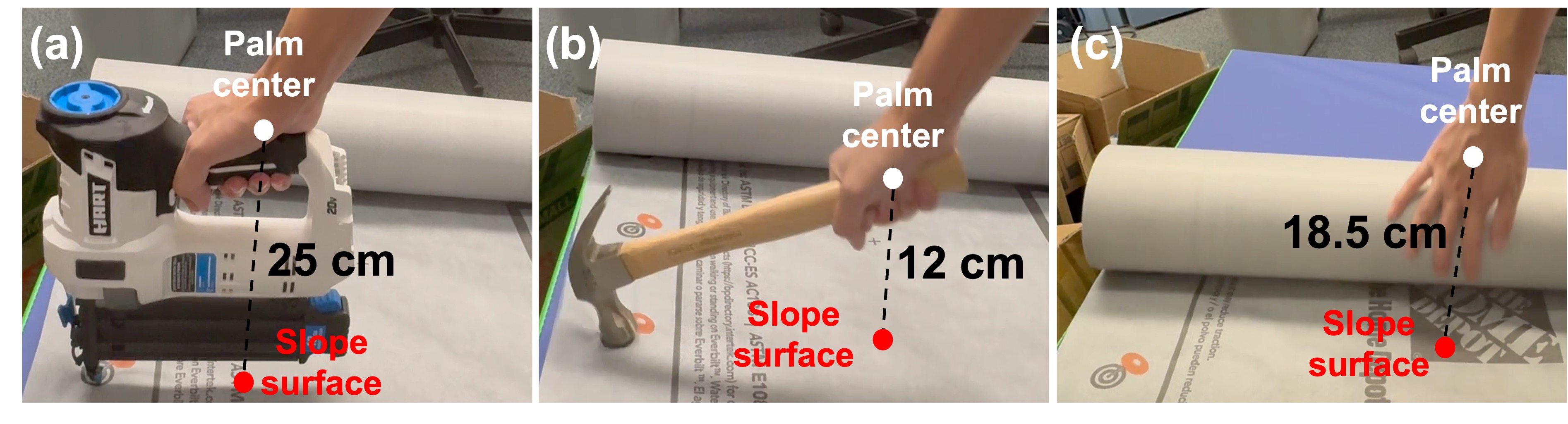}
    \caption{Measurement of task-specific hand-surface target clearances from human roofing demonstrations: (a) 25 cm for nailgun operation, (b) 12 cm for hammering, and (c) 18.5 cm for lateral material pushing.}
    \label{fig:target_gap}
\end{figure}

The robot is a 29-DoF Unitree G1. Policy learning uses $4096$ parallel environments, a $200$-Hz simulation rate with control decimation $4$ ($50$-Hz policy rate), and $30{,}000$ PPO iterations. Each iteration collects $24$ steps per environment and performs five epochs over four mini-batches. PPO uses discount $0.99$, generalized-advantage parameter $0.95$, clipping ratio $0.2$, entropy coefficient $0.005$, and adaptive learning rates initialized at $10^{-3}$.  The roof geometry is a hand-drawn metric mesh matched to the laboratory platform; the pushing task uses a two-module roof of twice the cross-slope width. Static and dynamic friction are randomized during training over $[0.3,1.6]$ and $[0.3,1.2]$, respectively. Deterministic evaluation disables pushes, pose noise, action delay, center-of-mass randomization, and friction variation. Fig.~\ref{fig:target_gap} shows our manual measurement of task-specific hand-surface target clearances from human roofing demonstrations. Each clearance is defined from the palm center to the roof surface along the local surface-normal direction during the intended work phase and is used as the target distance $d_w$ for reference refinement and execution-time policy learning.

\subsection{Evaluation Metrics}

For valid rollout frames $\mathcal{T}$, base- and world-frame MPJPE are defined as
\begin{equation}
E_{\mathrm{MPJPE}}
=
\frac{1}{|\mathcal{T}|J}
\sum_{t\in\mathcal{T}}
\sum_{j=1}^{J}
\left\|
\mathbf{x}^{\mathrm{exec}}_{j,t}
-
\mathbf{x}^{\mathrm{ref}}_{j,t}
\right\|_2 .
\end{equation}
In the simulation results, $\mathrm{MPJPE}_{b}$ denotes base-frame mean per-joint position error, and $\mathrm{MPJPE}_{w}$ denotes world-frame mean per-joint position error. The base-frame form removes global root translation and yaw and therefore measures local pose reproduction, whereas the world-frame form also includes global trajectory error. Both quantities are reported in millimeters, using the same notation in Tables~\ref{tab:motion_tracking} and~\ref{tab:slope_matrix}.

For work-task evaluation, $E_{\mathrm{foot}}$ is the mean absolute signed distance from active support points to the roof. The work-clearance error is
\begin{equation}
E_{\mathrm{work}}
=
\frac{1}{|\mathcal{T}_{w}|}
\sum_{t\in\mathcal{T}_{w}}
\left|
d_{\mathrm{hand},t}-d_w
\right|,
\end{equation}
where $\mathcal{T}_{w}$ is the detected work phase, $d_{\mathrm{hand},t}$ is the executed hand-center clearance, and $d_w$ is the task-specific desired clearance. We further report $d_{\mathrm{handmesh}}$, the minimum signed distance over sampled hand-mesh points during the work phase; a negative value indicates roof penetration. A rollout is counted as successful only if it completes the motion, covers the complete work phase, satisfies $E_{\mathrm{foot}}\leq2$ cm and $E_{\mathrm{work}}\leq3$ cm, and has $d_{\mathrm{handmesh}}\geq0$.

Because the two external baselines do not execute the same reference-conditioned control problem, we additionally report protocol-appropriate diagnostics. For reward-only learning, nominal episode completion is separated from roofing-task success. We compare the most favorable continuous equal-duration window with the recorded pushing prior after removing root translation and yaw, and report core-link position error, work-gap error, and lower-body joint-acceleration RMS. This best-window alignment favors the baseline and is interpreted as motion-prior deviation rather than synchronized tracking error. For SONIC, we report stable completion, fall occurrence, and phase-aligned local link-position error. These baseline-specific measurements are not pooled statistically with the A/M/B/C/D ablation.

The physical evaluation reports pelvis-frame MPJPE as mean $\pm$ sample standard deviation across five independent trials for each of nailgun positioning, hammering, and bending. The uphill-walking analysis additionally reports joint-angle RMSE and link-wise position errors. These measurements quantify local tracking fidelity, not unassisted fall probability or construction-task reliability. We do not report a comprehensive set of joint/link velocity and acceleration tracking errors or externally measured physical root-position and orientation errors. The absence of external localization and the use of a safety hoist limit claims about global trajectory accuracy and balance safety.

\subsection{Simulation Results}
Our simulation evaluation comprises five parts. First, Table~\ref{tab:motion_tracking} evaluates motion-tracking fidelity across a broad library of roof traversal, working-posture, and tool-related behaviors, establishing the behavioral coverage of the learned policies. Second, Table~\ref{tab:slope_matrix} examines how the roof pitches represented during training affect motion completion and tracking performance across $9^\circ$, $17^\circ$, and $25^\circ$ slopes. Third, Table~\ref{tab:nailgun_ablation} presents the complete A/M/B/C/D nailgun ablation, isolating the effects of raw retargeting, manual height correction, support grounding, task-semantic reference refinement, and execution-time semantic rewards. Fourth, Table~\ref{tab:cross_task} validates the proposed reference refinement and runtime enforcement on hammering and lateral pushing, which have different motion structures and desired hand--surface relations. Fifth, the reward-only RL and zero-shot SONIC teleoperation baselines examine whether roof-compatible roofer motions can be obtained without the proposed scene-grounded motion-prior pipeline.

\subsubsection{Motion-Coverage Evaluation}

We first evaluate the breadth of roofer-style behaviors represented by the learned policies. Table~\ref{tab:motion_tracking} summarizes the final simulation tracking performance across all evaluated motion categories. Across the traversal motions, the base-frame MPJPE is generally low, ranging from 18.35 mm to 37.07 mm, indicating that the learned policies can preserve the local body configuration of the reference motions. The \(9^\circ{+}17^\circ\) setting achieves the lowest base-frame error for uphill, downhill, lateral, and pivot motions, suggesting that moderate multi-slope training can improve local pose tracking for roof traversal. However, when the \(25^\circ\) slope is added, the error increases for most traversal motions, especially for uphill and downhill walking. For work postures and tool/object-related motions, the \(17^\circ\) setting often achieves the lowest world-frame MPJPE. This is reasonable because posture-dominant motions, such as stooping, kneeling, hammering, nailgun use, and pushing, require accurate local body placement near the roof surface rather than large root progression. Training on a single slope can therefore reduce ambiguity in terrain placement and produce lower global-frame errors. In contrast, adding more slope angles increases variation in the reference-to-terrain relationship, which can make global tracking more difficult for posture-rich motions.

\begin{figure}
    \centering
    \includegraphics[width=1\linewidth]{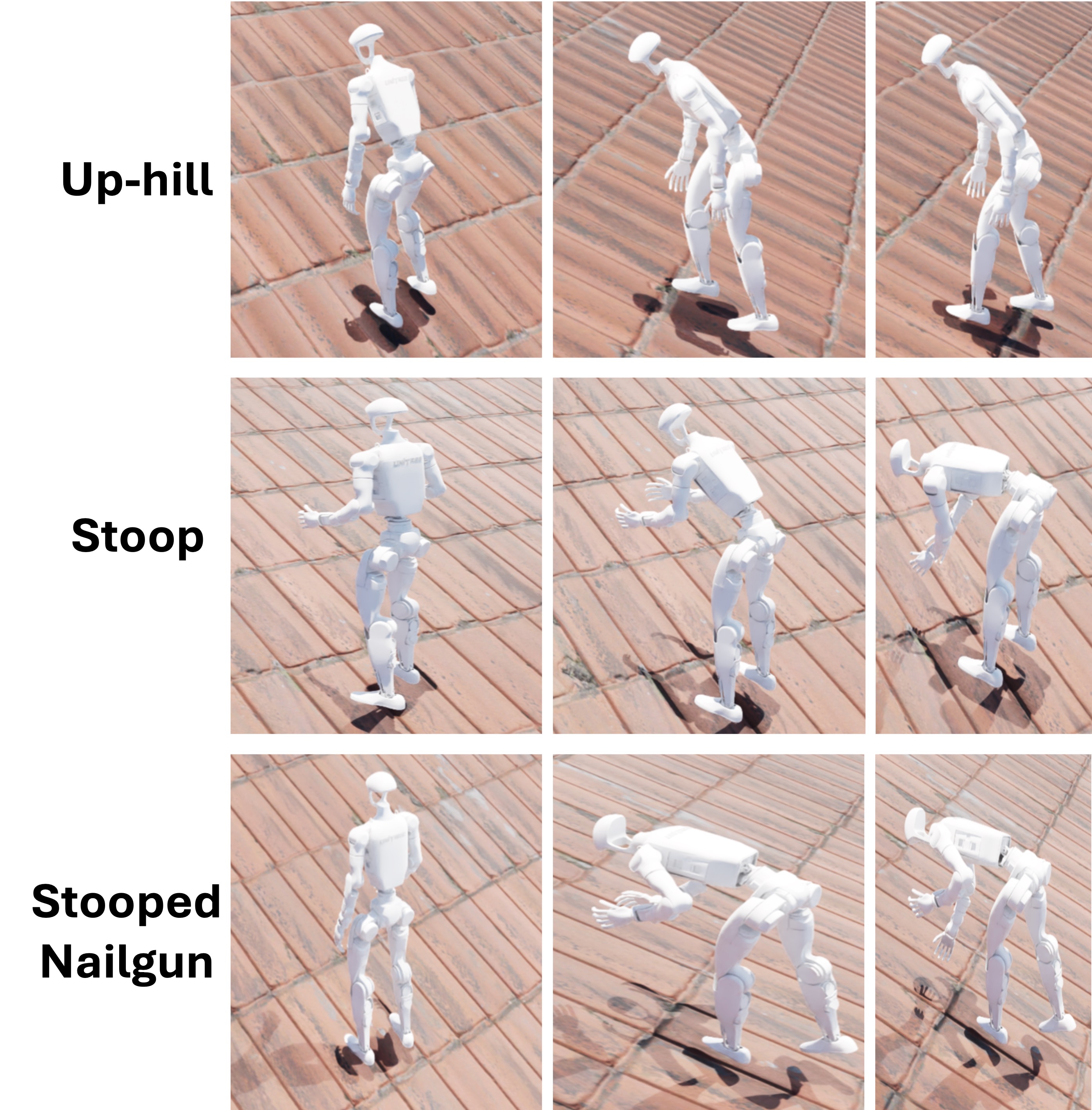}
    \caption{Examples of learned roofer-style motions in simulation, including uphill walking, stooping, and stooped nailgun operation on a sloped-roof scene.}
    \label{fig:8}
\end{figure}

Fig.~\ref{fig:8} visualizes representative learned motions in simulation, including uphill walking, stooping, and stooped nailgun operation. The snapshots show that the trained policy can reproduce both locomotion-dominant and posture-dominant behaviors on the sloped-roof scene. In uphill walking, the humanoid maintains forward progression while adapting its posture to the inclined surface. In stooping and stooped nailgun operation, the robot bends toward the roof while preserving a task-relevant whole-body configuration. These qualitative results complement the MPJPE results by showing that the learned policies do not merely minimize joint error, but also preserve the intended roofer-style motion structure.

\begin{table}[t]
\caption{Simulation tracking performance across the evaluated roofer-style motion library.}
\label{tab:motion_tracking}
\centering
\small
\renewcommand{\arraystretch}{1.15}
\begin{tabular}{c c c c c}
\hline
Motion group &
Motion &
Training slope set &
$\mathrm{MPJPE}_{b}$ (mm) &
$\mathrm{MPJPE}_{w}$ (mm) \\
\hline

\multirow{12}{*}{Traversal on roof}
& \multirow{3}{*}{Uphill}
& $17^\circ$ & 22.07 & 133.90 \\
& & $9^\circ{+}17^\circ$ & \textbf{20.62} & \textbf{128.20} \\
& & $9^\circ{+}17^\circ{+}25^\circ$ & 31.88 & 187.70 \\

& \multirow{3}{*}{Downhill}
& $17^\circ$ & 24.53 & \textbf{112.10} \\
& & $9^\circ{+}17^\circ$ & \textbf{23.46} & 128.90 \\
& & $9^\circ{+}17^\circ{+}25^\circ$ & 37.07 & 179.50 \\

& \multirow{3}{*}{Lateral}
& $17^\circ$ & 20.78 & \textbf{134.20} \\
& & $9^\circ{+}17^\circ$ & \textbf{20.76} & 141.80 \\
& & $9^\circ{+}17^\circ{+}25^\circ$ & 24.51 & 195.00 \\

& \multirow{3}{*}{Pivot}
& $17^\circ$ & 19.37 & 158.20 \\
& & $9^\circ{+}17^\circ$ & \textbf{18.35} & 155.10 \\
& & $9^\circ{+}17^\circ{+}25^\circ$ & 20.31 & \textbf{148.20} \\
\hline

\multirow{6}{*}{Work postures}
& \multirow{3}{*}{Stoop}
& $17^\circ$ & \textbf{28.26} & \textbf{146.90} \\
& & $9^\circ{+}17^\circ$ & 30.63 & 148.00 \\
& & $9^\circ{+}17^\circ{+}25^\circ$ & 31.40 & 176.90 \\

& \multirow{3}{*}{Kneeling}
& $17^\circ$ & 28.81 & \textbf{144.60} \\
& & $9^\circ{+}17^\circ$ & \textbf{28.00} & 164.60 \\
& & $9^\circ{+}17^\circ{+}25^\circ$ & 30.58 & 173.00 \\
\hline

\multirow{9}{*}{\begin{tabular}[c]{@{}c@{}}Tool/object-\\related motions\end{tabular}}
& \multirow{3}{*}{Stooped hammering}
& $17^\circ$ & \textbf{33.81} & \textbf{167.20} \\
& & $9^\circ{+}17^\circ$ & 37.94 & 173.00 \\
& & $9^\circ{+}17^\circ{+}25^\circ$ & 39.54 & 174.40 \\

& \multirow{3}{*}{Stooped nailgun}
& $17^\circ$ & \textbf{25.66} & \textbf{136.10} \\
& & $9^\circ{+}17^\circ$ & 26.26 & 146.10 \\
& & $9^\circ{+}17^\circ{+}25^\circ$ & 28.13 & 176.20 \\

& \multirow{3}{*}{Stooped pushing}
& $17^\circ$ & \textbf{26.14} & \textbf{102.20} \\
& & $9^\circ{+}17^\circ$ & 31.92 & 190.10 \\
& & $9^\circ{+}17^\circ{+}25^\circ$ & 37.55 & 207.70 \\
\hline
\multicolumn{5}{p{0.92\linewidth}}{Lower values indicate better tracking. Bold values indicate the lowest error for each motion and metric.} \\
\hline
\end{tabular}
\normalsize
\end{table}

\subsubsection{Slope-Coverage Evaluation}
This experiment examines whether trained RL policies can generalize to roof pitches outside those represented during training, and whether expanding the training-slope set improves robustness at steeper pitches. We compare policies trained on nested slope sets: $17^\circ$ only, $9^\circ+17^\circ$, and $9^\circ+17^\circ+25^\circ$. Each policy is evaluated on the same $9^\circ$, $17^\circ$, and $25^\circ$ slopes using the corresponding slope-matched uphill reference motion, ten deterministic episodes per policy--slope combination, and identical evaluation settings. MPJPE is computed only for successfully completed trajectories so that full rollouts are not compared with prematurely terminated ones. The nested training sets test the effect of progressively expanding slope coverage; this experiment is not intended to compare independent $9^\circ$-only and $25^\circ$-only specialist policies. As shown in Table~\ref{tab:slope_matrix}, the policy trained on $9^\circ+17^\circ$ completes the motions on both represented slopes but fails on the unseen $25^\circ$ slope. Similarly, the $17^\circ$-only policy fails all completion trials at $9^\circ$ and  $25^\circ$. After $25^\circ$ data are included during training, the policy achieves $100\%$ completion at all three evaluated pitches. These results indicate that depth input alone does not guarantee extrapolation to substantially steeper roofs; explicit slope coverage during training remains important. This is a bounded evaluation of the tested discrete pitches rather than evidence of continuous zero-shot slope generalization. Moreover, world-frame MPJPE increases with pitch even when completion remains high, particularly at $25^\circ$, indicating that successful motion completion can coexist with accumulated global root displacement.

\begin{table*}[t]
\caption{Policy-by-slope matrix for uphill walking. Each policy is evaluated on all three slopes using ten deterministic episodes per slope and identical evaluation settings. MPJPE values are mean $\pm$ sample standard deviation over successfully completed trajectories only. A dash indicates that no episode completed the full reference motion.}
\label{tab:slope_matrix}
\centering
\small
\setlength{\tabcolsep}{8pt}
\renewcommand{\arraystretch}{1.12}
\begin{tabular}{l c c c c}
\toprule
Training slopes &
Evaluation slope &
Success rate &
$\mathrm{MPJPE}_{b}$ (mm) &
$\mathrm{MPJPE}_{w}$ (mm) \\
\midrule

\multirow{3}{*}{$17^\circ$}
& $9^\circ$  & $0\%~(0/10)$   & ---                & --- \\
& $17^\circ$ & $100\%~(10/10)$ & $22.18\pm0.38$     & $93.94\pm16.69$ \\
& $25^\circ$ & $0\%~(0/10)$   & ---                & --- \\
\midrule

\multirow{3}{*}{$9^\circ+17^\circ$}
& $9^\circ$  & $100\%~(10/10)$ & $17.67\pm0.21$     & $60.49\pm3.18$ \\
& $17^\circ$ & $100\%~(10/10)$ & $20.95\pm0.34$     & $97.33\pm12.35$ \\
& $25^\circ$ & $0\%~(0/10)$   & ---                & --- \\
\midrule

\multirow{3}{*}{$9^\circ+17^\circ+25^\circ$}
& $9^\circ$  & $100\%~(10/10)$ & $17.97\pm0.51$     & $79.10\pm8.97$ \\
& $17^\circ$ & $100\%~(10/10)$ & $22.24\pm0.46$     & $96.75\pm8.94$ \\
& $25^\circ$ & $100\%~(10/10)$ & $23.74\pm0.27$     & $227.16\pm10.35$ \\
\bottomrule
\end{tabular}
\end{table*}


\subsubsection{Nailgun Ablation}

Table~\ref{tab:nailgun_ablation} examines three questions: whether raw retargeting produces a terrain-compatible reference, whether correcting foot support alone is sufficient for a roofing operation, and whether task-specific spatial relations must be enforced during policy execution. A uses the registered raw reference, M applies a constant vertical offset, and B introduces automatic support grounding. Although M and B substantially reduce foot error, neither represents the nailgun-specific hand--surface relation; consequently, the executed hand remains approximately $11$ cm above the roof instead of the desired $25$ cm and penetrates the roof mesh. C additionally incorporates the desired work clearance and hand-mesh nonpenetration into the offline reference refinement. However, its policy produces a mean hand gap of $32.86$ cm, demonstrating that a geometrically corrected reference does not guarantee that the relation is preserved under dynamic tracking. D therefore adds phase-gated execution-time semantic rewards. It reduces the work error to $0.531$ cm, maintains positive hand-mesh clearance, and succeeds for all three evaluated seeds. The ablation thus separates support grounding, task-semantic reference construction, and execution-time enforcement, showing that all three levels are needed for reliable task-surface interaction.

\begin{table*}[t]
\caption{Nailgun ablation over seeds 42, 123, and 456. Distances are mean $\pm$ sample standard deviation across seeds. The target gap from hand to slope surface is $25$ cm measured from human motions.}
\label{tab:nailgun_ablation}
\centering
\small
\setlength{\tabcolsep}{5pt}
\resizebox{\textwidth}{!}{%
\begin{tabular}{l c c c c c c}
\toprule
Method & Seeds &
$E_{\mathrm{foot}}$ (cm) &
$E_{\mathrm{work}}$ (cm) &
Mean hand gap (cm) &
$d_{\mathrm{handmesh}}$ (cm) &
Success\\
\midrule
A--Raw
& 3 & $4.837\pm2.049$ & $13.983\pm0.020$
& $11.017\pm0.020$ & $-1.945\pm0.270$ & $0/3$\\
M--Manual
& 3 & $0.447\pm0.084$ & $13.950\pm0.063$
& $11.050\pm0.063$ & $-2.054\pm0.172$ & $0/3$\\
B--Support
& 3 & $0.081\pm0.013$ & $13.973\pm0.027$
& $11.027\pm0.027$ & $-2.055\pm0.143$ & $0/3$\\
C--Reference
& 3 & $0.185\pm0.165$ & $7.860\pm0.328$
& $32.860\pm0.328$ & $19.198\pm0.099$ & $0/3$\\
D--Complete
& 3 & \textbf{$0.115\pm0.137$}
& \textbf{$0.531\pm0.104$}
& \textbf{$25.531\pm0.104$}
& $12.423\pm0.071$
& \textbf{$3/3$}\\
\bottomrule
\end{tabular}}
\end{table*}


\subsubsection{Cross-Task Validation}

After isolating the individual components through the nailgun ablation, we examine whether the benefit of execution-time task grounding persists across roofing operations with different motion structures and hand-surface requirements. Hammering represents a  close-surface operation with a desired hand-center clearance of $12$ cm, whereas pushing requires lateral whole-body movement while maintaining an $18.5$ cm clearance. Separate task-specific policies are trained for each motion; therefore, this experiment evaluates the consistency of the proposed method across tasks rather than zero-shot transfer of one policy.

Table~\ref{tab:cross_task} compares semantic reference correction alone (C) with the complete execution-aware method (D). For hammering, C achieves a relatively small work-clearance error of $1.105$ cm, but the minimum hand-mesh distance is $-1.772$ cm. This result shows that satisfying a hand-center target does not guarantee collision-free execution because another part of the hand can still penetrate the roof. For lateral pushing, C produces a mean hand gap of $16.245$ cm instead of the desired $18.5$ cm and also penetrates the roof by $1.792$ cm. In contrast, D reduces the work-clearance errors to $0.256$ cm for hammering and $0.424$ cm for pushing, while producing positive minimum hand-mesh clearances and succeeding across all three seeds for both tasks. These results demonstrate that scene-grounded reference correction provides a feasible task target, but execution-time clearance and nonpenetration rewards are necessary for preserving the intended spatial relation under dynamic policy-tracking errors.

\begin{table*}[t]
\caption{Cross-task comparison of semantic reference correction (C) and complete execution-aware learning (D).}
\label{tab:cross_task}
\centering
\small
\setlength{\tabcolsep}{4pt}
\resizebox{\textwidth}{!}{%
\begin{tabular}{l l c c c c c}
\toprule
Task (target gap) & Method &
$E_{\mathrm{foot}}$ (cm) &
$E_{\mathrm{work}}$ (cm) &
Mean hand gap (cm) &
$d_{\mathrm{handmesh}}$ (cm) &
Success\\
\midrule
\multirow{2}{*}{Hammer ($12$ cm)}
& C--Reference
& $0.539\pm0.057$
& $1.105\pm0.096$
& $11.030\pm0.122$
& $-1.772\pm0.236$
& $0/3$\\
& D--Complete
& \textbf{$0.492\pm0.036$}
& \textbf{$0.256\pm0.061$}
& \textbf{$11.940\pm0.087$}
& \textbf{$0.664\pm0.273$}
& \textbf{$3/3$}\\
\midrule
\multirow{2}{*}{Pushing ($18.5$ cm)}
& C--Reference
& $0.209\pm0.120$
& $3.254\pm0.567$
& $16.245\pm1.003$
& $-1.792\pm0.049$
& $0/3$\\
& D--Complete
& \textbf{$0.125\pm0.023$}
& \textbf{$0.424\pm0.222$}
& \textbf{$18.169\pm0.280$}
& \textbf{$4.528\pm0.299$}
& \textbf{$3/3$}\\
\bottomrule
\end{tabular}}
\end{table*}

\subsubsection{Pure RL and Zero-Shot SONIC Teleoperation Baselines}

\begin{figure}[t]
    \centering
    \includegraphics[width=\linewidth]{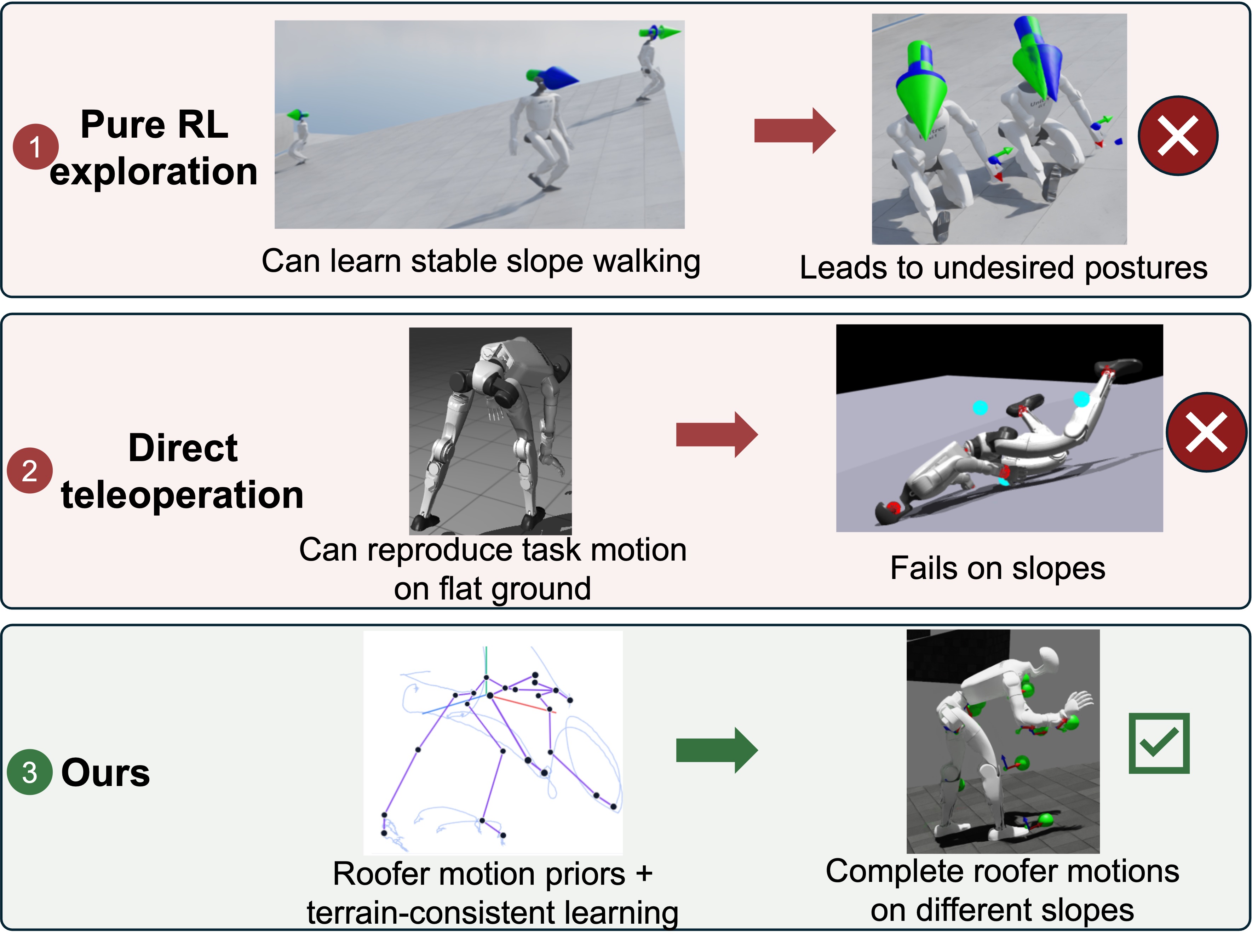}
    \caption{Representative failure modes of the external baselines. Reward-only learning produces a highly crouched motion that deviates from the human-derived pushing prior, whereas zero-shot SONIC teleoperation loses balance while stepping onto the roof.}
    \label{fig:qualitative_baselines}
\end{figure}

The two external baselines test whether the target behavior can be recovered without the proposed scene-grounded motion-learning pipeline. Representative failure modes are shown in Fig.~\ref{fig:qualitative_baselines}, and the quantitative results are summarized in Table~\ref{tab:baseline_comparison}.

The pure RL policy receives locomotion and hand-target rewards but no recorded roofer motion prior. It can move laterally on the $12^\circ$ slope, yet it discovers a highly crouched motion that differs substantially from the demonstrated pushing behavior, as illustrated in Fig.~\ref{fig:qualitative_baselines}. Across 100 evaluation episodes, 79 nominally complete the rollout, but only six satisfy the work-gap and nonpenetration success criteria. Even under favorable best-window alignment, its core-link error is $31.65\pm0.81$ cm, its work-gap error is $10.42\pm5.00$ cm, and its lower-body joint-acceleration RMS is $63.18\pm18.57$ rad/s$^2$, compared with $8.46$ rad/s$^2$ in the human-derived robot reference. These measurements do not define universal motion naturalness; instead, they show that task-reward completion alone does not recover the demonstrated roofing posture or its intended surface relation.

\begin{figure}[t]
    \centering
    \includegraphics[width=0.96\linewidth]{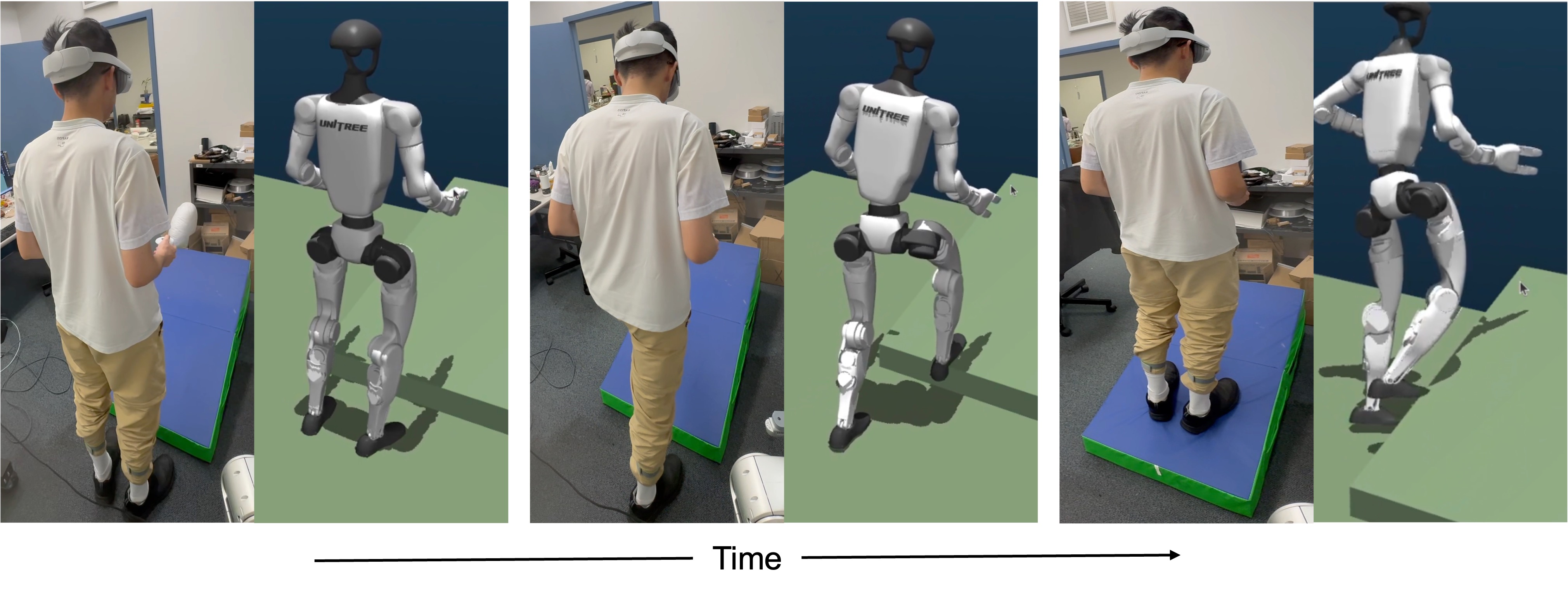}
    \caption{Zero-shot SONIC teleoperation failed to track the human motion of "stepping onto a slope".}
    \label{fig:sonic_failure}
\end{figure}

For the zero-shot teleoperation test, the same PICO operator performed two trials of stepping onto the physical roof platform with a nominal inclination of approximately $11.8^\circ$. The G1 in simulation, controlled by the open-source SONIC whole-body controller, tracked the operator and attempted to step onto a roof with the same nominal dimensions and inclination. As shown in Table~\ref{tab:baseline_comparison}, both recorded trials terminate after loss of balance or undesired terrain contact, yielding zero stable completions. Favorable phase-aligned local link-position errors are $63.4$ and $79.1$ mm for the two trials, while the corresponding initially aligned world-frame errors exceed $210$ mm. Fig.~\ref{fig:sonic_failure} shows the failure of zero-shot SONIC teleoperation to track the human motion of "stepping onto a ramp".
This result indicates that expressive whole-body teleoperation on benign support does not by itself provide roof-specific contact adaptation.

These external baselines are diagnostic comparisons rather than matched policy ablations. The reward-only policy has no motion reference, whereas SONIC uses online operator input and a different controller, observation interface, and objective. Their results are therefore not pooled with the proposed method or interpreted as a common performance ranking. The qualitative figures illustrate measured failure processes, not independent evidence of superiority. Evidence for the individual scene-grounding and execution-time components instead comes from the controlled A/M/B/C/D nailgun ablation and the three-seed C/D comparisons for hammering and pushing.

\begin{table*}[t]
\caption{Completion and roofing-task success of the external baselines.}
\label{tab:baseline_comparison}
\centering
\small
\setlength{\tabcolsep}{6pt}
\renewcommand{\arraystretch}{1.15}

\begin{tabular}{l c c c}
\toprule
Baseline &
\shortstack{Evaluation\\trials} &
\shortstack{Nominal/stable\\completion} &
\shortstack{Roofing-task\\success} \\
\midrule

Pure RL
& 100 episodes
& $79/100$
& $6/100$ \\

\shortstack[l]{SONIC}
& 2 trials
& $0/2$
& $0/2$ \\

\bottomrule
\end{tabular}
\end{table*}

\subsection{Physical-Robot Evaluation}

We further evaluate the simulation-trained policy on a physical Unitree G1 humanoid platform. The purpose of this experiment is to test whether the learned roofer-style motion can be transferred from simulation to a real robot under a sloped-roof setup. A safety hoist remains attached, and a human operator controls policy start and emergency stop. The physical roof platform is a Matladin folding gymnastics wedge mat with an expanded-polyethylene (EPE) foam core and a faux-leather outer surface. Its nominal width and horizontal run are 24 in ($0.6096$ m) and 48 in ($1.2192$ m), respectively, while its low- and high-edge heights are 4 in ($0.1016$ m) and 14 in ($0.3556$ m). The resulting rise is $0.2540$ m, corresponding to a nominal roof pitch of $\tan^{-1}(0.2540/1.2192)=11.768^\circ$.

The simulation mesh uses the same nominal platform dimensions and inclination. This correspondence does not establish equality of compliance, friction, or loaded surface shape. The overhead hoist is a fall-arrest safeguard and does not supply observations or control commands to the policy. Its presence nevertheless does not demonstrate that the robot remained unsupported throughout execution: hoist forces are not quantified, and the tether may restrict deep bending. Separate counts of uninterrupted completions, operator interventions, and hoist-assisted recoveries are not reported. Consequently, the repeated-trial MPJPE results are interpreted as preliminary, hoist-assisted motion-reproduction evidence rather than an estimate of unassisted completion probability or fall rate.

During execution, the deployment process records the measured joint angles from the G1 encoders and the corresponding time-indexed reference joint angles at 50 Hz; base orientation is also recorded. Using the same G1 URDF, forward kinematics is applied to both the measured and reference joint configurations to reconstruct the positions of 14 tracked links in the pelvis frame. Let $\mathbf p^{B}_{t,j}$ and $\mathbf p^{B,\mathrm{ref}}_{t,j}$ denote the reconstructed positions of link $j$ at frame $t$ for the measured and reference configurations, respectively. The frame-wise base-frame error is computed as
\[
e_t = \frac{1}{J}\sum_{j=1}^{J}
\left\|\mathbf p^{B}_{t,j}-\mathbf p^{B,\mathrm{ref}}_{t,j}\right\|_2,
\qquad J=14.
\]
The reported base-frame MPJPE is the mean of $e_t$ over all valid samples for which the reference motion is progressing:
\[
\mathrm{MPJPE}_{B} = \frac{1}{|\mathcal T|}\sum_{t\in\mathcal T}e_t,
\]
where $\mathcal T$ is the set of these valid samples; distances are reported in millimeters. This metric measures local pose tracking reconstructed from joint states rather than externally measured world-frame link positions. Because absolute root translation is unavailable from the onboard robot state and no external motion-capture or calibrated visual system is used, we do not report world-frame MPJPE or absolute robot-to-roof trajectory error.

\begin{figure}
    \centering
    \includegraphics[width=1\linewidth]{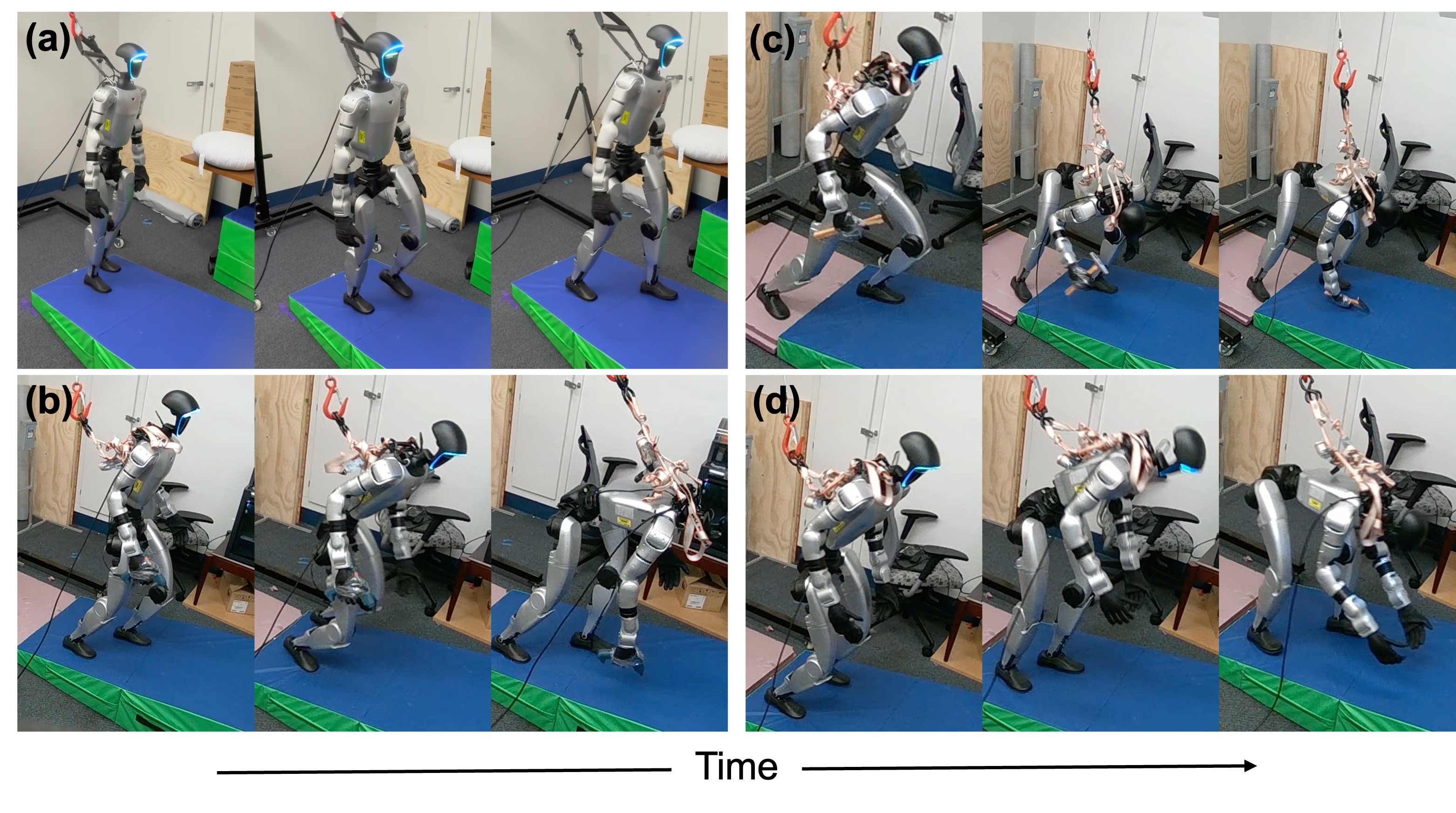}
    \caption{Real-robot execution of the learned four roofer motions on a slope. (a) uphill walking; (b) hold a nailgun and position it close to the slope surface; (c) hold and brandish a hammer; (d) bending without tools.}
    \label{fig:real-figure}
\end{figure}

Fig.~\ref{fig:real-figure}
shows the real-robot execution of the learned four roofer motions on a slope. Table~\ref{tab:real_robot} reports duration and measured base-frame MPJPE for four roofer motions executed by a physical G1 robot. Each of nailgun positioning, hammering, and bending is evaluated in five independent physical trials under the same roof geometry, initialization procedure, safety-hoist configuration, and policy-activation protocol. These motions add tool-related and working-posture validation to the uphill-walking traversal experiment. Nailgun, hammering, and bending produce mean base-frame MPJPE values of $60.2$, $79.9$, and $61.4$ mm over five trials, respectively. The larger errors than the uphill motion are consistent with deeper torso motion and greater arm excursion. These controlled tests provide preliminary evidence of deployment across the three motion groups without real-world policy optimization; they do not establish general reliability across operators, roof materials, or geometries.

\begin{table*}[t]
\caption{Recorded physical-G1 motion segments. Nailgun, hammering, and bending MPJPE values are mean $\pm$ sample standard deviation over five independent trials per motion; uphill walking reports one recorded trial.}
\label{tab:real_robot}
\centering
\small
\begin{tabular}{l c c}
\toprule
Motion & Duration (s) & Base MPJPE (mm)\\
\midrule
Uphill walking & 5.00 & 27.6\\
Nailgun  & 10.04 & $60.20\pm18.86$\\
Hammering  & 8.08 & $79.93\pm24.23$\\
Bending  & 6.56 & $61.42\pm19.09$\\
\bottomrule
\end{tabular}
\end{table*}

\begin{figure}
    \centering
    \includegraphics[width=1\linewidth]{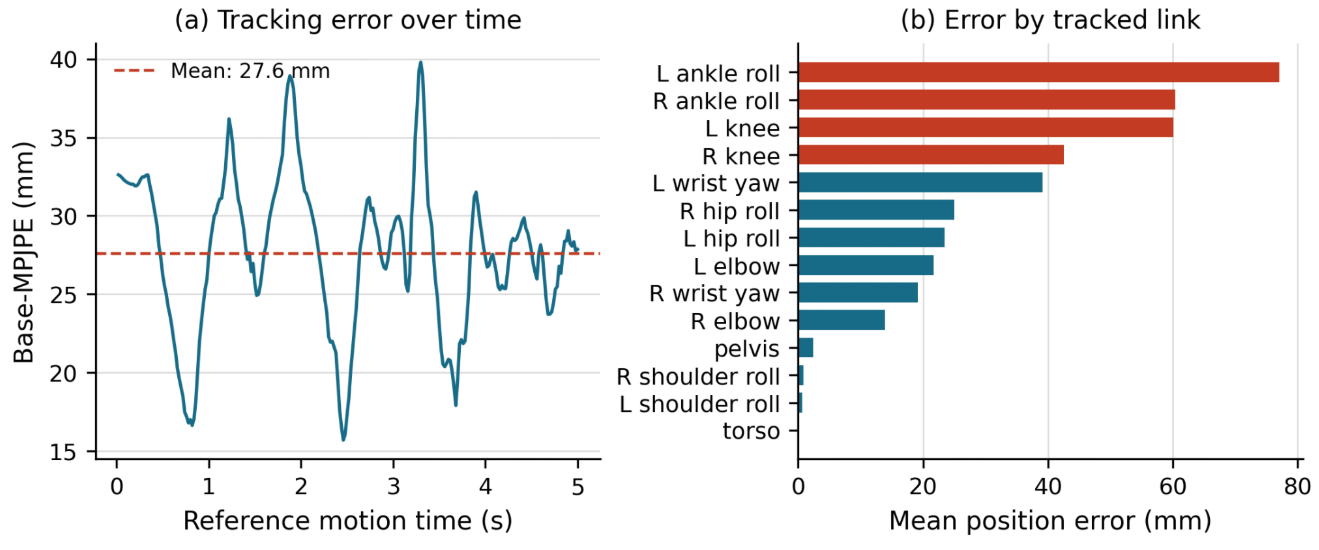}
    \caption{Real-robot tracking performance for the uphill-walking motion on a slope.}
    \label{fig:real-tracking}
\end{figure}

Fig.~\ref{fig:real-tracking} further analyzes the real-robot tracking error over time and across tracked links. The base-frame MPJPE remains within a moderate range during most of the execution, with several peaks occurring during phases where the robot changes support or adjusts its lower-body configuration. The joint-angle RMSE over all 29 joints was \(0.133~\mathrm{rad}\) \((7.61^\circ)\). The lower-body RMSE was \(0.164~\mathrm{rad}\) \((9.41^\circ)\), which is larger than the upper-body RMSE of \(0.088~\mathrm{rad}\) \((5.02^\circ)\). This difference suggests that the main real-world tracking difficulty comes from lower-body support and contact adaptation rather than upper-body motion tracking. The link-wise error distribution confirms this observation. The largest spatial tracking errors occur at the distal lower-body links, including \(87.17~\mathrm{mm}\) at the left ankle, \(65.90~\mathrm{mm}\) at the right ankle, \(69.53~\mathrm{mm}\) at the left knee, and \(46.37~\mathrm{mm}\) at the right knee. These errors are expected because uphill walking on a sloped surface requires continuous adjustment of the knees and ankles to maintain support and compensate for contact mismatch. In contrast, the upper-body links show smaller errors, indicating that the learned policy preserves the intended torso and arm motion more consistently.

\section{Discussion}
\subsection{Interpretation of Results}
The results support the central premise of this study: roofer-style humanoid behavior requires both a human-derived motion prior and explicit grounding to the roof surface. The motion-coverage evaluation shows that human demonstrations can provide references for traversal, low working postures, and tool-related motions. Base-frame MPJPE is generally lower than world-frame MPJPE, indicating that preserving local body posture does not necessarily ensure accurate global movement over the roof. Motion fidelity, trajectory fidelity, and task success must therefore be evaluated separately.

The slope-coverage results show that terrain observations alone do not guarantee extrapolation to substantially different roof pitches. Policies trained on $9^\circ+17^\circ$ slopes fail to complete the uphill motion on the unseen $25^\circ$ slope, whereas including $25^\circ$ data produces successful completion at all three evaluated pitches. Thus, explicit coverage of the intended slope range remains important. Because only three discrete pitches are evaluated, this result does not demonstrate continuous zero-shot generalization.

The nailgun ablation explains the roles of the proposed components. Rigid registration alone does not make a retargeted motion contact-consistent. Manual height correction and support grounding improve foot placement but cannot independently impose the required hand--surface relation. Reference-level semantic refinement produces a geometrically meaningful target, but policy-tracking errors can still violate it during execution. Adding phase-gated clearance and nonpenetration rewards allows D to maintain the desired work relation and succeed across all three seeds. The hammering and pushing experiments show the same trend under different motion structures and target clearances. In particular, the hammering result demonstrates that a correct hand-center distance is insufficient when another part of the hand mesh can still penetrate the roof.

The external baselines further motivate the proposed formulation. Pure RL frequently completes the nominal episode but rarely satisfies the complete roofing-task criteria and produces motions that deviate substantially from the human-derived pushing prior. Zero-shot SONIC teleoperation fails in both trials when the robot attempts to step onto the slope. They show that task rewards alone may admit unintended motion strategies, while direct human-motion teleoperation does not automatically provide roof-specific contact adaptation.

The physical experiments provide preliminary evidence that simulation-trained policies can reproduce uphill walking, nailgun, hammering, and bending motions on the laboratory G1 without real-world policy optimization. Tracking errors are larger for the work motions than for uphill walking, and the largest discrepancies occur around the knees and ankles. This suggests that lower-body support adaptation remains the main sim-to-real challenge. Because only onboard joint states and base orientation are recorded, the current physical evaluation measures base-frame motion fidelity but cannot directly verify world-frame trajectory, foot placement, or hand-roof clearance.

\subsection{Implications for Construction Engineering}

Many construction activities are defined by spatial relations with a work surface rather than by motion alone. Examples include maintaining a tool at an appropriate clearance, aligning material with a roof plane, or permitting knee contact only during a planned phase. The proposed framework represents these relations explicitly instead of relying only on pose imitation or generic locomotion rewards. The division between human demonstration and scene geometry is also compatible with construction workflows. Low-cost wearable tracking provides worker motion, coordination, and task timing, while a measured, designed, or scanned roof model provides metric surface geometry. Such models could be obtained from building information models or site scans and used to refine robot motions before deployment. The resulting policies should be viewed as motion primitives that could later be combined with perception, foothold selection, task planning, tool control, and safety monitoring in a more complete construction system.

The evaluation also suggests that construction-robot performance should not be measured only through joint or pose error. Support accuracy, tool clearance, body-surface penetration, task-phase coverage, and task success provide a more direct connection to construction requirements. Future evaluations could extend these metrics to fastening accuracy, material alignment, applied force, surface coverage, and installation quality.

\subsection{Limitations and Future Work}

The experiments use controlled laboratory platforms represented by predominantly planar meshes. Real roofs contain shingles, seams, ridges, debris, damaged regions, compliance, and spatially varying friction. Although friction is randomized in simulation, robustness to scanned-mesh noise, surface uncertainty, roof edges, and weather conditions has not been established. Future work should incorporate local surface estimation, uncertainty-aware safety margins, and testing on more diverse roof materials and geometries.

The semantic specification is also only partially automatic. Foot-support phases are inferred from motion cues, but work phases and target clearances rely on task annotation and measurements from human operator. The palm or controller-derived point is used as a proxy for the hand/tool location, and the full tool and material geometries are not modeled. Consequently, the experiments evaluate work-related whole-body positioning rather than functional outcomes such as nail firing, hammer impact, pushing force, or installation quality. Future systems should model tool geometry explicitly and infer task phases and tolerances from multiple demonstrations.

Finally, the physical experiments use a safety hoist, manually controlled policy activation, and a small number of trials to prevent hardware damage. The hoist may keep the robot from finishing deep bending motions, and the absence of external robot-to-roof tracking prevents direct measurement of world-frame and task-surface errors. Future experiments should use calibrated external cameras or motion capture, characterize hoist forces, report more repeated task-level success rates, and evaluate slip recovery, roof-edge avoidance, and emergency stopping before considering autonomous construction deployment.

\section{Conclusion}

This study investigates how human roofing demonstrations can be converted into dynamically executable humanoid motions on pitched surfaces. The results show that preserving human motion alone is insufficient: support placement and task-specific work relations must also be grounded in metric roof geometry and maintained during policy execution. The main conclusions are as follows:

\begin{itemize}

\item Scene grounding resolves geometric errors that cannot be addressed by direct retargeting or a uniform height offset. The proposed formulation independently constrains foot support, task-effector clearance, and body-mesh nonpenetration while preserving the characteristic whole-body structure of the demonstrated roofing motion.

\item The simulation ablations demonstrate that both reference-level correction and execution-time enforcement are necessary. Support correction improves foot placement but does not recover the intended hand-surface relation, while a feasible semantic reference can still be violated by policy-tracking errors. The complete method achieves work-clearance errors of $0.531\pm0.104$ cm for nailgun use, $0.256\pm0.061$ cm for hammering, and $0.424\pm0.222$ cm for lateral pushing, with positive hand-mesh clearance and successful execution across all three evaluated seeds for each task.

\item The broader evaluation shows that the learned policies reproduce multiple traversal, posture, and tool-related motion primitives, but robustness remains dependent on the training distribution. Including the $25^\circ$ slope during training enables successful completion at all three evaluated pitches, whereas policies without this slope fail at $25^\circ$. The pure RL and zero-shot SONIC results further indicate that generic task rewards or direct teleoperation do not reliably recover both roofer-style motion and roof-specific contact adaptation.

\item Physical experiments provide preliminary evidence that the simulation-trained policies can reproduce uphill walking, nailgun, hammering, and bending motions on a laboratory Unitree G1 without real-world policy optimization. The recorded base-frame MPJPE ranges from $27.6$ to $79.9$ mm.

\end{itemize}

\section{Data Availability Statement}
Some or all data, models, or code that support the findings of this research are available from the corresponding author upon reasonable request.

\section{Acknowledgments}
This research was supported by the U.S. National Science Foundation (Nos. 2222810). Any
opinions, findings, and conclusions or recommendations
expressed in this paper are those of the authors and do
not necessarily reflect the views of the National Science
Foundation, the University of Florida.

\section{Author Contributions}
Songyang Liu:  Investigation; Methodology; Visualization; Writing – original draft; Writing – review and editing. Shuai Li: Conceptualization; Supervision; Writing – review and editing.

\bibliography{ascexmpl-new}

@article{peng2018deepmimic,
  title={Deepmimic: Example-guided deep reinforcement learning of physics-based character skills},
  author={Peng, Xue Bin and Abbeel, Pieter and Levine, Sergey and Van de Panne, Michiel},
  journal={ACM Transactions On Graphics (TOG)},
  volume={37},
  number={4},
  pages={1--14},
  year={2018},
  publisher={ACM New York, NY, USA}
}

@inproceedings{luo2023perpetual,
  title={Perpetual humanoid control for real-time simulated avatars},
  author={Luo, Zhengyi and Cao, Jinkun and Winkler, Alexander and Kitani, Kris and Xu, Weipeng},
  booktitle={Proceedings of the IEEE/CVF International Conference on Computer Vision},
  pages={10895--10904},
  year={2023}
}

@inproceedings{he2024omnih2o,
  title     = {{OmniH2O}: Universal and Dexterous Human-to-Humanoid Whole-Body Teleoperation and Learning},
  author    = {He, Tairan and Luo, Zhengyi and He, Xialin and Xiao, Wenli and Zhang, Chong and Zhang, Weinan and Kitani, Kris M. and Liu, Changliu and Shi, Guanya},
  booktitle = {Proceedings of the 8th Conference on Robot Learning},
  series    = {Proceedings of Machine Learning Research},
  volume    = {270},
  pages     = {1516--1540},
  publisher = {PMLR},
  year      = {2025},
  url       = {https://proceedings.mlr.press/v270/he25b.html}
}

@inproceedings{he2025asap,
  title     = {{ASAP}: Aligning Simulation and Real-World Physics for Learning Agile Humanoid Whole-Body Skills},
  author    = {He, Tairan and Gao, Jiawei and Xiao, Wenli and Zhang, Yuanhang and Wang, Zi and Wang, Jiashun and Luo, Zhengyi and He, Guanqi and Sobanbabu, Nikhil and Pan, Chaoyi and Yi, Zeji and Qu, Guannan and Kitani, Kris and Hodgins, Jessica K. and Fan, Linxi and Zhu, Yuke and Liu, Changliu and Shi, Guanya},
  booktitle = {Proceedings of Robotics: Science and Systems},
  address   = {Los Angeles, CA},
  year      = {2025},
  doi       = {10.15607/RSS.2025.XXI.066}
}

@inproceedings{yang2025omniretarget,
  title     = {{OmniRetarget}: Interaction-Preserving Data Generation for Humanoid Whole-Body Loco-Manipulation and Scene Interaction},
  author    = {Yang, Lujie and Huang, Xiaoyu and Wu, Zhen and Kanazawa, Angjoo and Abbeel, Pieter and Sferrazza, Carmelo and Liu, C. Karen and Duan, Rocky and Shi, Guanya},
  booktitle = {Proceedings of the IEEE International Conference on Robotics and Automation (ICRA)},
  year      = {2026},
  publisher = {IEEE},
  url       = {https://omniretarget.github.io/}
}

@misc{iko2023minimumslope,
  author       = {{IKO}},
  title        = {What is the Minimum Slope for an Asphalt Shingle Roof},
  year         = {2023},
  month        = nov,
  day          = {22},
  howpublished = {\url{https://www.iko.com/na/blog/minimum-slope-for-asphalt-shingle-roof/}},
  note         = {Accessed: 2026-06-28}
}

@article{zhuang2026deep,
  title={Deep Whole-body Parkour},
  author={Zhuang, Ziwen and Zhu, Shaoting and Zhao, Mengjie and Zhao, Hang},
  journal={arXiv preprint arXiv:2601.07701},
  year={2026}
}

@inproceedings{ivaldi2023telemovtop,
  author    = {Ivaldi, Serena and Ghini, Edoardo},
  title     = {Teleoperating a Robot for Removing Asbestos Tiles on Roofs: Insights from a Pilot Study},
  booktitle = {2023 IEEE International Conference on Advanced Robotics and Its Social Impacts (ARSO)},
  pages     = {128--133},
  year      = {2023},
  doi       = {10.1109/ARSO56563.2023.10187476}
}

@misc{bdr2026roofus,
  author       = {{Building Diagnostic Robotics}},
  title        = {Roofus Roof Inspection Platform: Technical Specifications},
  year         = {2026},
  howpublished = {\url{https://buildingdiagnosticrobotics.com/roofus-tech-specs}},
  note         = {Accessed: 2026-06-10}
}

@misc{renovate2026rufus,
  author       = {{Renovate Robotics}},
  title        = {Renovate Robotics},
  year         = {2026},
  howpublished = {\url{https://www.renovaterobotics.com/}},
  note         = {Accessed: 2026-06-10}
}

@inproceedings{hezaveh2017roofdamage,
  author    = {Mahdavi Hezaveh, Mahshad and Kanan, Christopher and Salvaggio, Carl},
  title     = {Roof Damage Assessment Using Deep Learning},
  booktitle = {2017 IEEE Applied Imagery Pattern Recognition Workshop (AIPR)},
  year      = {2017},
  doi       = {10.1109/AIPR.2017.8457946}
}

@article{xu2022roofdamage,
  author  = {Xu, Jinglin and Zeng, Feng and Liu, Wen and Takahashi, Toru},
  title   = {Damage Detection and Level Classification of Roof Damage after Typhoon Faxai Based on Aerial Photos and Deep Learning},
  journal = {Applied Sciences},
  volume  = {12},
  number  = {10},
  pages   = {4912},
  year    = {2022},
  doi     = {10.3390/app12104912}
}

@article{kucharczyk2025poststormroof,
  author  = {Kucharczyk, Michal},
  title   = {Automated Mapping of Post-Storm Roof Damage Using Deep Learning and Aerial Imagery: A Case Study in the Caribbean},
  journal = {Remote Sensing},
  volume  = {17},
  number  = {20},
  pages   = {3456},
  year    = {2025},
  doi     = {10.3390/rs17203456}
}

@article{zahradnik2023flatroof,
  author  = {Zahradn{\'i}k, David and Rou{\v{c}}ka, Filip and Karlovsk{\'a}, Linda},
  title   = {Flat Roof Classification and Leaks Detections by Deep Learning},
  journal = {Stavebn{\'i} obzor -- Civil Engineering Journal},
  year    = {2023},
  doi     = {10.14311/CEJ.2023.04.0042}
}

@article{mostafa2023builtuproof,
  author  = {Mostafa, Kareem and Hegazy, Tarek and Hunsperger, Robert D. and Elias, Stepanka},
  title   = {Using Image Analysis to Quantify Defects and Prioritize Repairs in Built-Up Roofs},
  journal = {Facilities},
  volume  = {41},
  number  = {7/8},
  pages   = {498--509},
  year    = {2023},
  doi     = {10.1108/F-08-2022-0119}
}

@article{santos2023flatroofequipment,
  author  = {Santos, Lara Monalisa Alves dos and Zanoni, Vanda Alice Garcia and Bedin, Eduardo and Pistori, Hemerson},
  title   = {Deep Learning Applied to Equipment Detection on Flat Roofs in Images Captured by UAV},
  journal = {Case Studies in Construction Materials},
  volume  = {18},
  pages   = {e01917},
  year    = {2023},
  doi     = {10.1016/j.cscm.2023.e01917}
}

@inproceedings{li2024roofsubcomponents,
  author    = {Li, Jiajun and Tao, Boan and Bosch{\'e}, Fr{\'e}d{\'e}ric and Lu, Chris Xiaoxuan and Wilson, Lyn},
  title     = {Extracting roof sub-components from orthophotos using deep-learning-based semantic segmentation},
  booktitle = {Proceedings of the 41st International Symposium on Automation and Robotics in Construction},
  pages     = {675--682},
  year      = {2024},
  doi       = {10.22260/ISARC2024/0088}
}

@article{alzarrad2022roofaiuav,
  author  = {Alzarrad, Ammar and Awolusi, Ibukun and Hatamleh, Muhammad T. and Terreno, Saratu},
  title   = {Automatic assessment of roofs conditions using artificial intelligence {(AI)} and unmanned aerial vehicles {(UAVs)}},
  journal = {Frontiers in Built Environment},
  volume  = {8},
  pages   = {1026225},
  year    = {2022},
  doi     = {10.3389/fbuil.2022.1026225}
}

@article{zhao2025rrdsegnet,
  author  = {Zhao, Xiayu and Jebelli, Houtan},
  title   = {A computational method for real-time roof defect segmentation in robotic inspection},
  journal = {Computer-Aided Civil and Infrastructure Engineering},
  volume  = {40},
  number  = {23},
  pages   = {3596--3623},
  year    = {2025},
  doi     = {10.1111/mice.13471}
}

@article{zhao2025saha,
  author  = {Zhao, Xiayu and Liu, Yizhi and Jebelli, Houtan},
  title   = {Module-enhanced slope-adaptive and hazard-aware hexapod robotic system for safe roof inspection},
  journal = {ASCE OPEN: Multidisciplinary Journal of Civil Engineering},
  volume  = {3},
  number  = {1},
  pages   = {04025011},
  year    = {2025},
  doi     = {10.1061/AOMJAH.AOENG-0088}
}

@inproceedings{romano2021nailedit,
  author    = {Romano, Matthew M. and Chen, Yuxin and Kuevor, Prince and Marshall, Owen and Atkins, Ella M.},
  title     = {Nailed It: Autonomous Roofing with a Nailgun-Equipped Octocopter},
  booktitle = {AIAA AVIATION 2021 FORUM},
  pages     = {3211},
  year      = {2021},
  doi       = {10.2514/6.2021-3211}
}

@article{li2024roofdt,
  title   = {Automated generation and semantic segmentation of roof orthophoto for digital twin-based monitoring of slated roofs},
  author  = {Li, Jiajun and Tao, Boan and Bosché, Frédéric and Lu, Chris Xiaoxuan and Wilson, Lyn},
  journal = {Automation in Construction},
  volume  = {167},
  pages   = {105725},
  year    = {2024},
  doi     = {10.1016/j.autcon.2024.105725}
}

@article{radosavovic2024challengingterrain,
  title         = {Learning Humanoid Locomotion over Challenging Terrain},
  author        = {Radosavovic, Ilija and Kamat, Sarthak and Darrell, Trevor and Malik, Jitendra},
  journal       = {arXiv preprint arXiv:2410.03654},
  year          = {2024},
  doi           = {10.48550/arXiv.2410.03654},
  archivePrefix = {arXiv},
  eprint        = {2410.03654},
  primaryClass  = {cs.RO}
}

@article{kuindersma2016atlas,
  title   = {Optimization-based locomotion planning, estimation, and control design for the {Atlas} humanoid robot},
  author  = {Kuindersma, Scott and Deits, Robin and Fallon, Maurice and Valenzuela, Andr{\'e}s and Dai, Hongkai and Permenter, Frank and Koolen, Twan and Marion, Pat and Tedrake, Russ},
  journal = {Autonomous Robots},
  volume  = {40},
  number  = {3},
  pages   = {429--455},
  year    = {2016},
  doi     = {10.1007/s10514-015-9479-3}
}

@article{radosavovic2024realworld,
  title   = {Real-world humanoid locomotion with reinforcement learning},
  author  = {Radosavovic, Ilija and Xiao, Tete and Zhang, Bike and Darrell, Trevor and Malik, Jitendra and Sreenath, Koushil},
  journal = {Science Robotics},
  volume  = {9},
  number  = {89},
  pages   = {eadi9579},
  year    = {2024},
  doi     = {10.1126/scirobotics.adi9579}
}

@inproceedings{zhuang2025parkour,
  title     = {Humanoid Parkour Learning},
  author    = {Zhuang, Ziwen and Yao, Shenzhe and Zhao, Hang},
  booktitle = {Proceedings of the 8th Conference on Robot Learning},
  series    = {Proceedings of Machine Learning Research},
  volume    = {270},
  pages     = {1975--1991},
  publisher = {PMLR},
  year      = {2025},
  url       = {https://proceedings.mlr.press/v270/zhuang25a.html}
}

@inproceedings{wang2025beamdojo,
  title     = {{BeamDojo}: Learning Agile Humanoid Locomotion on Sparse Footholds},
  author    = {Wang, Huayi and Wang, Zirui and Ren, Junli and Ben, Qingwei and Huang, Tao and Zhang, Weinan and Pang, Jiangmiao},
  booktitle = {Proceedings of Robotics: Science and Systems},
  address   = {Los Angeles, CA},
  year      = {2025},
  doi       = {10.15607/RSS.2025.XXI.068}
}

@inproceedings{gu2024dwl,
  title     = {Advancing Humanoid Locomotion: Mastering Challenging Terrains with Denoising World Model Learning},
  author    = {Gu, Xinyang and Wang, Yen-Jen and Zhu, Xiang and Shi, Chengming and Guo, Yanjiang and Liu, Yichen and Chen, Jianyu},
  booktitle = {Proceedings of Robotics: Science and Systems},
  address   = {Delft, Netherlands},
  year      = {2024},
  doi       = {10.15607/RSS.2024.XX.058}
}

@inproceedings{sun2025perceptive,
  title     = {Learning Perceptive Humanoid Locomotion over Challenging Terrain},
  author    = {Sun, Wandong and Cao, Baoshi and Chen, Long and Su, Yongbo and Liu, Yang and Xie, Zongwu and Liu, Hong},
  booktitle = {Proceedings of the IEEE/RSJ International Conference on Intelligent Robots and Systems (IROS)},
  pages     = {6571--6578},
  publisher = {IEEE},
  year      = {2025},
  doi       = {10.1109/IROS60139.2025.11247685}
}

@inproceedings{chen2025gmt,
  title     = {{GMT}: General Motion Tracking for Humanoid Whole-Body Control},
  author    = {Chen, Zixuan and Ji, Mazeyu and Cheng, Xuxin and Peng, Xuanbin and Peng, Xue Bin and Wang, Xiaolong},
  booktitle = {IEEE/RSJ International Conference on Intelligent Robots and Systems (IROS)},
  year      = {2026},
  note      = {Accepted for publication},
  url       = {https://gmt-humanoid.github.io/}
}

@article{luo2025sonic,
  title   = {{SONIC}: Supersizing motion tracking for natural humanoid whole-body control},
  author  = {Luo, Zhengyi and Yuan, Ye and Wang, Tingwu and Li, Chenran and Casta{\~n}eda, Fernando and Chen, Sirui and Cao, Zi-Ang and Li, Jiefeng and Minor, David and Ben, Qingwei and Park, Jinhyung and Sami, David and Wang, Zi and Da, Xingye and Ding, Runyu and Hogg, Cyrus and Song, Lina and Lim, Edy and Jeong, Eugene and He, Tairan and Xue, Haoru and Xiao, Wenli and Yuen, Simon and Kautz, Jan and Chang, Yan and Iqbal, Umar and Fan, Linxi Jim and Zhu, Yuke},
  journal = {Science Robotics},
  volume  = {11},
  number  = {117},
  pages   = {eaed4592},
  year    = {2026},
  doi     = {10.1126/scirobotics.aed4592}
}

@inproceedings{araujo2025retargeting,
  title     = {Retargeting Matters: General Motion Retargeting for Humanoid Motion Tracking},
  author    = {Ara{\'u}jo, Jo{\~a}o Pedro and Ze, Yanjie and Xu, Pei and Wu, Jiajun and Liu, C. Karen},
  booktitle = {Proceedings of the IEEE International Conference on Robotics and Automation (ICRA)},
  year      = {2026},
  publisher = {IEEE},
  url       = {\url{https://jiajunwu.com/papers/gmr_icra.pdf}}
}

@misc{osha_fall_workbook,
  title        = {Managing Fall Protection Hazards Workbook},
  author       = {{Occupational Safety and Health Administration}},
  year         = {2015},
  howpublished = {\url{https://www.osha.gov/sites/default/files/2018-12/fy15_sh-27683-sh5_Fall_Prevention_Student_Workbook_English.pdf}},
  note         = {Accessed 2026-03-08}
}

@misc{osha_roofing_workers,
  author = {{Occupational Safety and Health Administration}},
  title  = {Protecting Roofing Workers},
  year   = {2015},
  note   = {Available online: https://www.osha.gov/sites/default/files/publications/OSHA3755.pdf, accessed 2026-03-08}
}

@misc{cdc_falls_2019,
  title        = {Prevent Construction Falls from Roofs, Ladders, and Scaffolds},
  author       = {{National Institute for Occupational Safety and Health}},
  year         = {2019},
  howpublished = {\url{https://www.cdc.gov/niosh/docs/2019-128/pdfs/2019-128Revised112019.pdf}},
  note         = {Accessed 2026-03-08}
}

@misc{bls_fatal_falls_2023,
  title        = {Fatal Falls in the Construction Industry in 2023},
  author       = {{U.S. Bureau of Labor Statistics}},
  year         = {2025},
  howpublished = {\url{https://www.bls.gov/opub/ted/2025/fatal-falls-in-the-construction-industry-in-2023.htm}},
  note         = {Accessed 2026-03-08}
}

@misc{cpwr_falls_2024,
  title        = {Data Bulletin: Falls, Slips, and Trips in Construction},
  author       = {{CPWR -- The Center for Construction Research and Training}},
  year         = {2024},
  howpublished = {\url{https://www.cpwr.com/wp-content/uploads/DataBulletin-March2024.pdf}},
  note         = {Accessed 2026-03-08}
}

@article{liao2026beyondmimic,
  title   = {{BeyondMimic}: From Motion Tracking to Versatile Humanoid Control via Guided Diffusion},
  author  = {Liao, Qiayuan and Truong, Takara E. and Huang, Xiaoyu and Gao, Yuman and Tevet, Guy and Sreenath, Koushil and Liu, C. Karen},
  journal = {Science Robotics},
  year    = {2026},
  doi     = {10.1126/scirobotics.adx8924}
}

\end{document}